\documentclass[10pt,twocolumn,letterpaper]{article}

\usepackage{cvpr}
\PassOptionsToPackage{table}{xcolor}
\usepackage[T1]{fontenc}
\usepackage{times}
\usepackage{graphicx}
\usepackage{booktabs}
\usepackage{amsmath,amssymb}
\usepackage{array}
\usepackage{tabularx}
\usepackage{pifont}
\usepackage[table]{xcolor}
\usepackage{url}
\usepackage{placeins}
\usepackage{microtype}

\renewcommand{\arraystretch}{1.05}
\newcommand{\benchmark}{Seeing What Execution Needs}

\newcommand{\ledger}{Spatial--Functional Belief Ledger}
\newcommand{\aibr}{Action-Induced Belief Revision}

\definecolor{tosblue}{RGB}{232,240,252}
\definecolor{tosorange}{RGB}{254,242,232}
\definecolor{tosgray}{RGB}{245,245,245}

\definecolor{cvprblue}{rgb}{0.21,0.49,0.74}
\usepackage[pagebackref,breaklinks,colorlinks,allcolors=cvprblue]{hyperref}

\def\confName{CVPR}
\def\confYear{2027}

\title{Seeing What Execution Needs: Benchmarking Task-Conditioned Actionable Spatial Belief for Manipulation}
\author{
Yina Jian$^{1,2}$ \qquad Mingkun Xu$^{3,\dagger}$\\
$^{1}$Columbia University \quad $^{2}$Stanford University\\
$^{3}$Guangdong Institute of Intelligence Science and Technology, Zhuhai, China\\
{\tt\small yj2713@columbia.edu} \qquad
{\tt\small yinajan@stanford.edu} \qquad
{\tt\small xumingkun@gdiist.cn}\\
{\small $^\dagger$Corresponding author}
}

\begin{document}
\maketitle

\begin{abstract}
When should a robot look at the cup, and when at the portafilter?
A view that is informative for one task can be useless for another,
even at the same workstation.
We introduce \benchmark{}, a benchmark for task-conditioned
actionable spatial belief in simulated tabletop manipulation.
Its aligned three-domain track contains 1{,}080 frozen cases
(360 each for coffee workstation, drawer/cabinet, and
microwave/appliance).
The primary coffee protocols---closed-loop actionability and
same-object functional out-of-distribution generalization---produce
over 115{,}000 evaluator episodes across methods, budgets, and
inference repeats.
Each case pairs candidate RGB views with a manipulation command,
shared scripted skills, and hidden postconditions, directly
measuring whether selected observations support execution.
A metadata-assisted diagnostic reference pipeline maintains
spatial--functional beliefs and ranks views by missing evidence.
At primary budget settings, it reaches 98.8\% closed-loop and
98.9\% functional-OOD success; removing functional fields or
command binding reduces functional-OOD success to 0.0\% and
19.7\%, respectively.
Under a separate full-preview contract, seven vision-language
models reach 26--43\%.
Replacing coverage metadata with frozen RGB-derived scores
retains 98.5\%/98.9\% success; shuffling these scores reduces
success to 15\%/11\%, supporting dependence on image content.
Separately, the aligned three-domain evaluation at $H{=}3$
supports portability of the benchmark interface under frozen
domain-specific schemas.
These results suggest that spatial belief for manipulation should
be evaluated by whether acquired evidence meets the current
command's execution requirements.
\end{abstract}

\begin{figure}[!t]
    \centering
    \includegraphics[width=0.50\columnwidth]{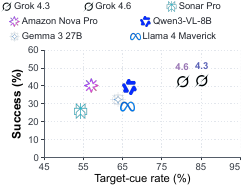}\hfill
    \includegraphics[width=0.50\columnwidth]{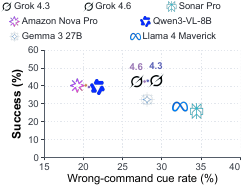}
    \caption{\textbf{Seven-model evidence diagnostics.}
    Left: success versus target-cue rate.
    Right: success versus wrong-command cue rate.
    Seven strict-clean VLMs each have 1{,}500 evaluation records,
    preview twelve RGB candidates, and select exactly three views.
    Dots show point estimates; icons label models.
    Cue rates are separate evaluator diagnostics, not complementary
    fractions.}
    \label{fig:strict-vlm-cue-scatter}
    \vspace{-8pt}
\end{figure}
\begin{figure*}[t]
    \centering
    \includegraphics[width=0.93\textwidth]{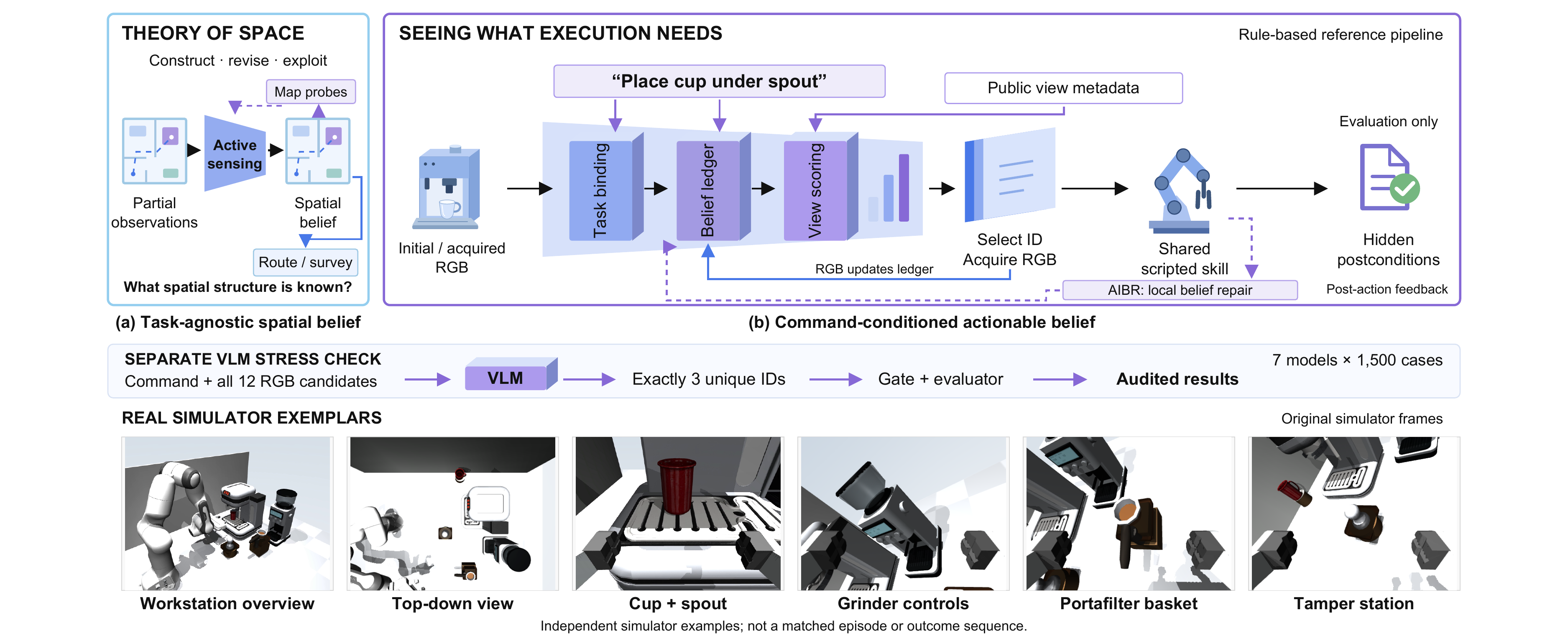}
    \caption{\textbf{From spatial belief to actionable evidence.}
    Theory-of-Space-style evaluation asks whether an agent can
    construct and revise spatial beliefs through exploration.
    \benchmark{} keeps the active-belief motivation but scores
    whether command-conditioned views expose the evidence needed
    by a shared scripted skill, with hidden postconditions used
    only for evaluation.
    Schematic modules are illustrative.}
    \label{fig:geometric-overview}
    \vspace{-8pt}
\end{figure*}

\section{Introduction}
An embodied agent can look at the right scene and still acquire
the wrong evidence.
Consider a coffee workstation: tamping requires observing the
portafilter basket, puck surface, and tamper alignment; cup
placement requires cup--spout alignment, support, and clearance;
grinding requires the hopper, chute, receiver, and relevant
controls.
The physical workspace may be unchanged, yet the observation
that makes one command executable can be irrelevant to another.
Seeing more is not necessarily seeing what execution needs.

Recent work has expanded spatial reasoning from passive question
answering toward active belief construction and embodied
interaction~\citep{das2018embodiedqa,wang2025enact}.
Theory of Space studies whether agents can acquire, maintain,
and revise spatial beliefs through
exploration~\citep{zhang2026theoryspace};
ESI-Bench evaluates spatial intelligence through
perception--action loops~\citep{hong2026esibench};
and BEHAVIOR-Spatial connects spatial understanding and
instruction following to downstream robot
behavior~\citep{chen2026behaviorspatial}.
In parallel, vision-language-action policies and
affordance-centered methods increasingly ground language in
object parts, action regions, and manipulation
trajectories~\citep{brohan2023rt2,kim2024openvla,
qu2025spatialvla,mo2021where2act,yu2025seqafford},
while active-vision systems learn where to look before
acting~\citep{bajcsy1988active,aloimonos1988active,
xiong2025visioninaction,kerr2025eyerobot,liu2026sapave,
shi2025visograsp,wang2026observeractor}.

These directions motivate a narrower but consequential question:
\emph{given a command, can an agent acquire the particular
spatial-functional evidence required to execute it?}
Existing evaluations may measure the accuracy of a spatial
answer, the quality of a global representation, or the final
success of a learned policy.
Final success alone, however, does not reveal whether an agent
selected the correct evidence, relied on a shortcut, or was
rescued by the controller.
Conversely, a geometrically accurate belief may contain many
facts while omitting the one relation that makes the next action
safe and executable.
A controlled evaluation should therefore hold the scene,
candidate observations, and downstream skill fixed, vary the
command and observation budget, and test whether the acquired
evidence supports a hidden task outcome.

\begin{figure*}[t]
\begin{minipage}[t]{\dimexpr(\textwidth-\columnsep)/2\relax}
    \vspace{0pt}
    \centering
    \includegraphics[width=0.93\columnwidth]{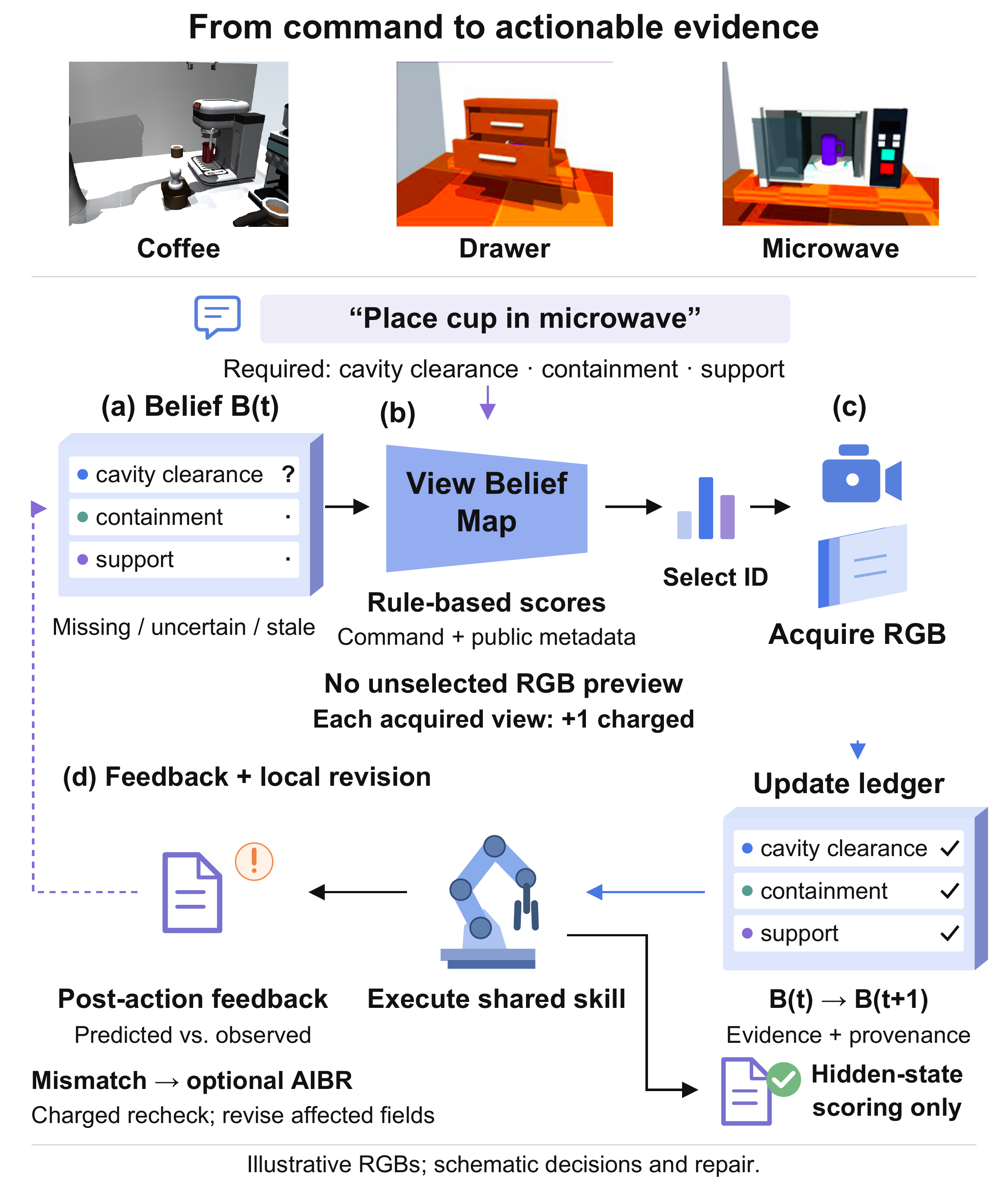}
    \caption{\textbf{Task-conditioned evidence acquisition.}
    The coffee workstation supplies the quantitative five-task
    benchmark; Drawer/Cabinet and Microwave/Appliance are evaluated at equal scale (360 cases each); full three-domain results appear in the supplement.
    (a)~The command and ledger identify missing evidence.
    (b)~The transparent View Belief Map ranks candidate views.
    (c)~Acquired observations update the ledger before a shared
    scripted skill executes.
    (d)~Post-action observations support optional local revision;
    hidden simulator state is reserved for evaluation.
    This hybrid schematic combines abstract icons with real MuJoCo
    RGB exemplars; the exemplars are illustrative interface views,
    not recorded selections or outcome evidence.}
    \label{fig:method-overview}
\end{minipage}\hfill
\begin{minipage}[t]{\dimexpr(\textwidth-\columnsep)/2\relax}
    \vspace{0pt}
    \centering
    \includegraphics[width=0.93\columnwidth]{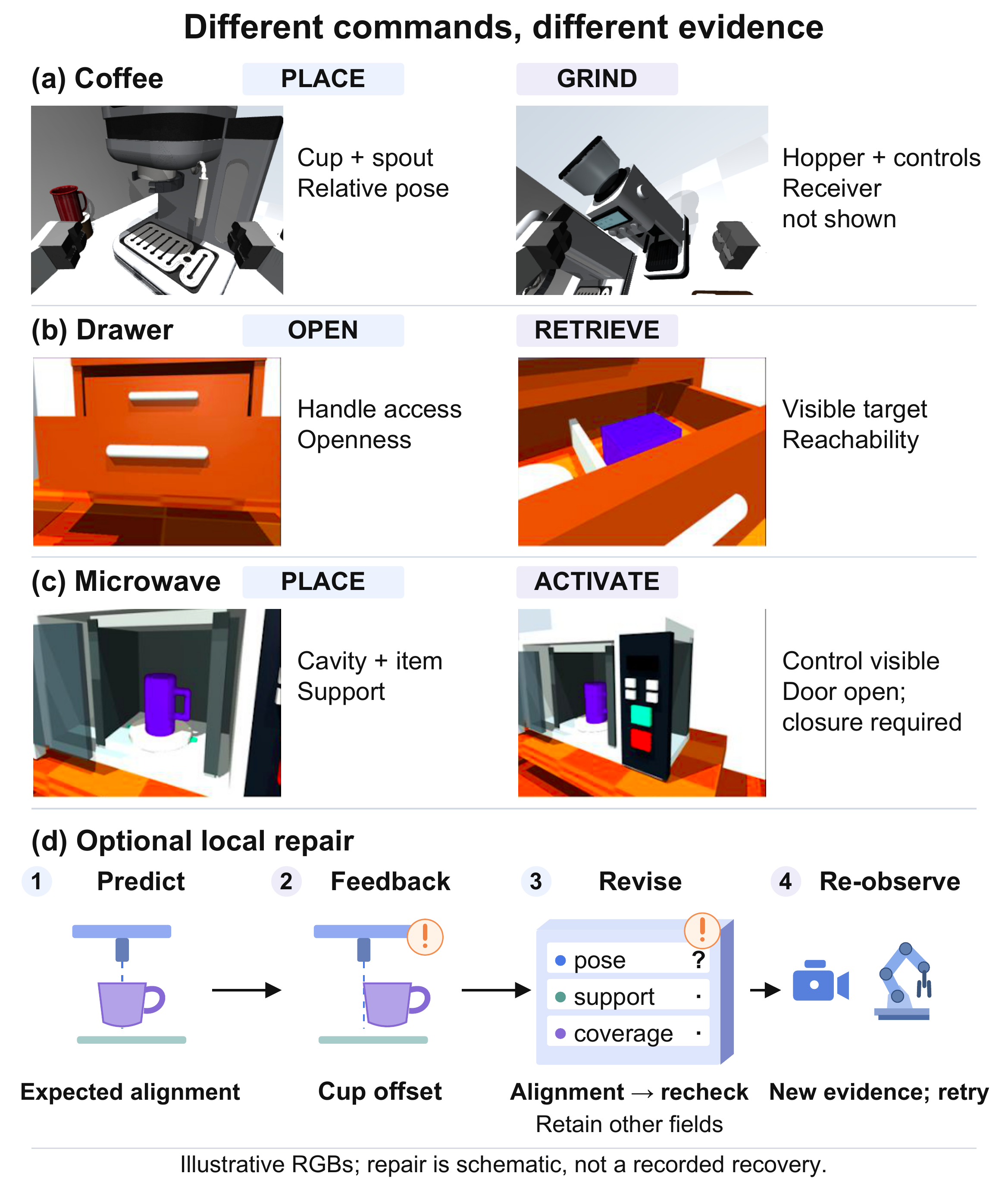}
    \caption{\textbf{Commands determine useful evidence; mismatches
    motivate local revision.}
    (a)~Coffee placement requires cup--spout alignment, whereas
    grinding requires the hopper--chute--basket chain.
    (b)~Drawer opening and retrieval emphasize handle access,
    openness, target visibility, and reachability.
    (c)~Microwave placement and activation emphasize cavity
    support and door/control state.
    (d)~An illustrative coffee-placement mismatch triggers
    rechecking of alignment while retaining support evidence,
    followed by re-observation and retry.
    This hybrid schematic combines abstract annotations with real
    MuJoCo RGB exemplars; highlighted regions are not measured
    detections, and the repair example does not establish
    cross-domain recovery performance.}
    \label{fig:command-functional-evidence}
\end{minipage}
\end{figure*}

We introduce \emph{Seeing What Execution Needs}, a frozen RGB benchmark
for \emph{task-conditioned actionable spatial belief}
(\cref{fig:geometric-overview}).
The benchmark spans three simulated tabletop domains:
a five-task coffee workstation (500~primary cases covering tamping, portafilter insertion, cup placement, serving, and grinding),
a drawer/cabinet interface (360~cases emphasizing handle access,
openness, and reachability), and a microwave/appliance interface
(360~cases emphasizing cavity support and control
state)---1{,}080~evaluation cases in the unified three-domain track
(\cref{fig:method-overview,fig:command-functional-evidence}).
Each case pairs a command with frozen candidate-view RGB
observations (twelve per coffee case, ten per drawer and
microwave case), a shared scripted skill interface, and hidden
simulator-state postconditions; the resulting base contains
11{,}520 candidate-view records.
Expansion across evaluated methods, observation budgets, and
repeated runs produces more than 115{,}000 evaluator episodes,
including 60{,}000 closed-loop and 55{,}500 same-object
functional-OOD episodes from the primary coffee domain.
The functional-OOD episodes preserve object identity while
changing command-conditioned functional requirements, testing
whether a selector follows the requested task rather than an
object- or scene-level shortcut.


The evaluation separates two sensing contracts.
Under the primary sequential protocol, every policy starts from
the same frozen candidate set and initial state, acquires
observations under a common budget~$H$, and passes the resulting
evidence to the same scripted skill.
Under a separate strict-clean VLM stress test, a model previews
all twelve frozen RGB candidates and must return exactly three
ranked view identifiers; this second experiment evaluates
full-candidate visual ranking rather than an equivalent
active-acquisition policy.
Both protocols reserve simulator-state postconditions for
evaluation: hidden state is never supplied to the view selector.
This separation makes it possible to attribute failures to
evidence acquisition without conflating them with learned
low-level control.

We additionally provide a transparent, metadata-assisted
reference pipeline.
A Spatial--Functional Belief Ledger records task-relevant pose,
relation, function, provenance, confidence, and staleness fields.
A rule-based View Belief Map binds the current command to
required predicates and ranks candidate views by their expected
gain over missing evidence; it uses the command, public
candidate-view metadata, and the current ledger state, but not
hidden postconditions or labels from unselected observations.
Optional Action-Induced Belief Revision compares predicted and
observed postconditions after execution, localizes potentially
stale ledger entries, and requests targeted re-observation.
The purpose of this pipeline is not to establish a general
learned vision policy but to provide an auditable reference
whose command-to-evidence decisions can be inspected directly.

The results expose a substantial gap between task-conditioned
evidence selection and generic visual ranking.
At the primary settings, the diagnostic reference pipeline reaches
$98.8\%$ closed-loop success at $H\!=\!5$ and $98.9\%$
functional-OOD success at $H\!=\!3$.
Removing functional belief fields reduces functional-OOD
success to $0.0\%$; removing command binding reduces it
to $19.7\%$.
Under the more restrictive $H\!=\!1$ initial-selection setting,
closed-loop and functional-OOD success fall to $19.0\%$ and
$58.6\%$, respectively.
In the separate full-candidate stress test, seven strict-clean
VLMs achieve only $26.0\%$--$42.9\%$ postcondition success
despite previewing every candidate
image~(\cref{fig:strict-vlm-cue-scatter}): models frequently
detect target-related cues without consistently selecting
evidence that supports the hidden task outcome.
These findings suggest that actionable belief requires more than
scene recognition: it requires binding the current command to
the evidence needed for execution.

\paragraph{Contributions.}
Our contributions are threefold:
\begin{itemize}
    \item We formulate \emph{task-conditioned actionable
      spatial-functional belief}, evaluating a spatial belief by
      whether it exposes the parts, relations, and functional
      states required by the current command.
      We introduce a frozen RGB benchmark spanning three
      simulated tabletop domains (five-task coffee workstation,
      drawer/cabinet, and microwave/appliance) with 1{,}080~evaluation cases, 11{,}520 candidate-view records, shared scripted skills, hidden simulator-state postconditions, and controlled closed-loop and functional-OOD evaluation.
    \item We provide an auditable metadata-assisted reference
      pipeline comprising a Spatial--Functional Belief Ledger,
      a rule-based View Belief Map, and optional Action-Induced
      Belief Revision, while explicitly separating public
      selection information from hidden evaluator state.
    \item We evaluate budget-matched baselines, ablations,
      no-image and shortcut controls, paired scene-group
      bootstrap estimates, observation-cost behavior, and seven
      strict-clean VLMs, accompanied by manifests, raw records,
      table-generation artifacts, and traceable admission checks.
\end{itemize}

\Cref{fig:geometric-overview} summarizes the central distinction:
spatial-belief evaluation asks what an agent knows about a scene,
whereas \emph{Seeing What Execution Needs} asks whether that belief helps the agent acquire the evidence made actionable by the current command.

\section{Related Work}

\paragraph{Spatial intelligence and embodied evaluation.}
Spatial-reasoning evaluation has advanced across complementary
layers.
Benchmarks now test spatial memory and measurement from video,
egocentric embodied cognition, and reference frames and spatial
relations in 2D/3D
observations~\citep{yang2025thinking,dang2025ecbench,
song2025robospatial};
diagnostic and comprehensive evaluation suites probe
spatial-reasoning
breadth~\citep{wang2025spatial457,wu2026spatialscore};
and vision-language methods train spatial reasoning, optimize 3D
layouts, and support agentic spatial
understanding~\citep{chen2024spatialvlm,sun2025layoutvlm,
marsili2025vadar}.
EmbodiedQA and Habitat established goal-directed navigation and
interactive evaluation in simulated environments, while ENACT
evaluates egocentric interaction through world
modeling~\citep{das2018embodiedqa,savva2019habitat,wang2025enact}.
XVR trains cross-view correspondence, verification, and
localization and transfers the resulting representations to
manipulation~\citep{jeong2026xvr}.
Theory of Space asks whether an exploring agent forms and revises
an internal spatial belief, connecting modern foundation-model
evaluation to classical planning under partial
observability~\citep{zhang2026theoryspace,kaelbling1998planning}.
These works principally score answers, representations, or broad
agent behavior.
We focus on a narrower controlled interface: whether observations
selected \emph{before execution} expose the command-specific
evidence required for a shared manipulation skill to satisfy a
hidden postcondition.

\paragraph{Manipulation benchmarks and policies.}
RLBench, ManiSkill2, and CALVIN provide standardized task suites
for learning and evaluating diverse or long-horizon manipulation
policies~\citep{james2020rlbench,gu2023maniskill2,mees2022calvin};
CLIPort and VIMA further test language- or multimodal-prompt-conditioned
tabletop
generalization~\citep{shridhar2022cliport,jiang2023vima}.
ALFRED and TEACh evaluate language-grounded household task
execution, with TEACh additionally requiring interactive
dialogue~\citep{shridhar2020alfred,padmakumar2022teach}.
SAPIEN, Habitat, and BEHAVIOR-1K provide complementary simulation
infrastructure and activity suites for controlled embodied
evaluation~\citep{xiang2020sapien,savva2019habitat,li2024behavior1k}.
These settings primarily vary the learned policy and score
end-to-end task completion or generalization.
Our benchmark instead fixes the downstream skill and hidden
evaluator, isolating whether a method first acquires the
command-specific spatial-functional evidence that makes execution
possible; this decomposition complements, rather than replaces,
evaluation on broad manipulation suites.
Within the learned-policy literature, Where2Act localizes
actionable regions on articulated objects, SeqAfford grounds
sequential 3D affordances, and RT-2, OpenVLA, SpatialVLA, and
$\pi_0$ map visual-language context to robot actions or spatially
grounded action
representations~\citep{mo2021where2act,yu2025seqafford,
brohan2023rt2,kim2024openvla,qu2025spatialvla,black2025pi0}.
ActiveVLA and SaPaVe explicitly incorporate critical-region
localization or active view--pose
reasoning~\citep{liu2026activevla,liu2026sapave}.
PhysVLM focuses on physical reachability, while RoboGround
grounds manipulation with vision-language
priors~\citep{zhou2025physvlm,huang2025roboground}.
GraspGPT and ThinkGrasp use language-model or vision-language
reasoning for task-oriented grasping and strategic part grasping
in clutter~\citep{tang2023graspgpt,qian2025thinkgrasp}.

\paragraph{Active perception for embodied action.}
Classic active-vision and next-best-view work selects sensing
actions to reduce uncertainty or improve
reconstruction~\citep{bajcsy1988active,aloimonos1988active,
connolly1985nbv,pito1999nbv,scott2003viewplanning}.
Recent robotic systems learn task-directed gaze, camera motion,
or base placement: Vision in Action learns human-like search and
focus from demonstrations; EyeRobot couples gaze with hand
actions; Observe Then Act coordinates camera next-best-view
selection with gripper next-best-pose planning; and Mobi-$\pi$
chooses robot base poses that make a downstream policy
effective~\citep{xiong2025visioninaction,kerr2025eyerobot,
wang2025observe,yang2025mobipi}.
VISO-Grasp and Observer-Actor further couple active view planning
or sparse-view perception to
manipulation~\citep{shi2025visograsp,wang2026observeractor}.
WoMAP likewise uses learned world models for active
open-vocabulary localization under partial
observability~\citep{yin2025womap}.
TVVE~\cite{bai2026tvve} learns a task-aware virtual-viewpoint
exploration policy in a pseudo-environment;
MAE-Select~\cite{yi2026viewpoint} trains a single-camera
active-viewpoint selector without viewpoint labels;
and INSPECT~\cite{wen2026inspect} learns robot view
preferences from records of a smart-glasses assistant, explicitly
separating evidence requirements from camera-reproducible
observation changes.
These works learn a viewpoint-selection policy; our benchmark
fixes the candidate set and downstream skill, isolating
whether a selector acquires the right evidence under a
controlled evaluation contract.

\paragraph{Closest concurrent benchmarks.}
FUSE studies learned semantic--geometric evidence acquisition for
functional affordance grounding, ESI-Bench broadens embodied
spatial evaluation to perception--action loops, VABench studies
visual demonstrations, active perception, and metric control, and
BEHAVIOR-Spatial connects spatial understanding and instruction
following to robot
behavior~\citep{chen2026fuse,hong2026esibench,
zhang2026vabench,chen2026behaviorspatial}.
These concurrent efforts broaden the learned-policy or agent
setting.
\benchmark{} contributes a controlled complement: frozen RGB
candidates, explicit sensing contracts, command-shift
interventions, a fixed skill interface, hidden simulator-state
postconditions, and a metadata-assisted reference selector whose
decision trace can be audited.
This narrower design is intended to isolate \emph{which evidence
was acquired and why}, not to establish a general-purpose
active-vision policy.

\section{Benchmark and Method}

\paragraph{Task interface.}
Each episode contains a hidden simulator state~$s$, a manipulation command~$\tau$, a set of candidate views~$\mathcal{V}$, and RGB observations acquired under budget~$H$. The agent sees the command, acquired RGB frames, candidate-view metadata, and its own ledger state; it never sees hidden predicates, unselected endpoint renders, evaluator labels, contact logs, or postcondition ground truth before acting. The five task families---tamp, insert, place, serve, and grind---each have 50~scene groups and two variants in the primary coffee domain, yielding 500~frozen coffee cases and 6{,}000~unique candidate images. The aligned cross-domain track uses a 360-case subset of Coffee together with 360 Drawer/Cabinet and 360 Microwave/Appliance cases, yielding 1{,}080~cases and 11{,}520~candidate images. Coffee provides 12 candidates per case, whereas each held-out domain provides 10; the acquisition budget is fixed to $H{=}3$ throughout this track. All downstream execution uses the same shared scripted skill library, so differences in task outcome trace to evidence acquisition rather than controller quality.

\paragraph{Evaluation protocols.}
The benchmark defines two protocols that isolate different aspects of
evidence acquisition.
Protocol~C.1 (\emph{closed-loop actionability}, primary budget
$H{=}5$) tests sequential view selection: the agent acquires up to
five views one at a time, updating its ledger after each observation,
then passes the accumulated evidence to the scripted skill.
Protocol~C.2 (\emph{same-object functional OOD}, $H{=}3$) holds
object identity fixed across commands and varies only the functional
requirements---a cup that must be \emph{placed} needs alignment and
clearance evidence, whereas the same cup in a \emph{serve} command
needs fill-level and stability evidence.
This split exposes whether a selector follows the commanded task or
falls back on an object- or scene-level shortcut.
In both protocols, hidden simulator-state postconditions are
evaluated only after execution: the evaluator checks whether the
physical outcome satisfies the command's requirements without
revealing those conditions to the agent beforehand.

\paragraph{Spatial--functional belief ledger.}
The \ledger{} stores entries
\[
B_t=\{(p_i,\hat y_i,q_i,E_i,\kappa_i,\rho_i)\}_{i=1}^{N_t},
\]
where $p_i$ is a spatial or functional predicate, $\hat y_i$ is its
estimated value, $q_i$ is confidence, $E_i$ records visual
provenance, $\kappa_i$ marks staleness, and $\rho_i$ marks task
relevance.
Predicates include part visibility, relative pose, support,
containment, action affordance, functional cue, and predicted
postcondition state.
In the reference implementation, $\rho_i$ is instantiated from
command templates, $E_i$ stores the selected view IDs that support
an entry, $q_i$ is a three-level confidence flag (observed, inferred,
or unknown), and $\kappa_i$ is raised only by actions that may
invalidate the corresponding predicate type.
The benchmark evaluates whether a policy acquires the command-relevant
evidence it needs; the ledger's role is to make that evaluation
transparent and traceable, not to serve as an optimal learned state
estimator.

\paragraph{View Belief Map.}
For each candidate view~$v$, the agent estimates which missing
target predicates the view may resolve:
\[
J(v;B_t,\tau)=
\frac{1}{|\mathcal M_t|}\sum_{p_i\in\mathcal M_t}
g(v,p_i,\tau)-\lambda\,c(v),
\]
where $\mathcal M_t$ is the set of missing, uncertain, or stale
target-relevant predicates, $g$ is an ordinal evidence-gain estimate,
and $c(v)$ is normalized view cost.
The selected views form a discrete View Belief Map over the candidate
camera set---a command-conditioned score over executable observations,
not a dense point-cloud value field.

\begin{figure}[t]
    \centering
    \includegraphics[width=\linewidth,trim=36bp 3bp 33bp 4bp,clip]{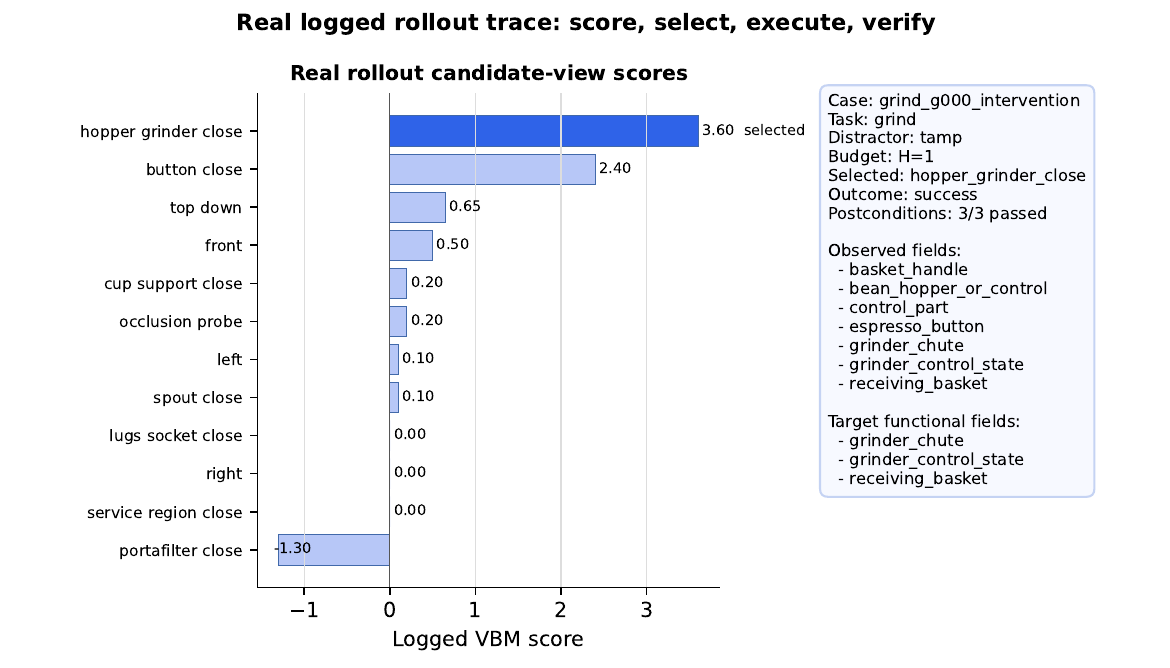}
    \caption{\textbf{Logged rollout trace for command-conditioned view
    selection.}
    This panel is generated from a real frozen functional-OOD trace.
    It shows the candidate-view scores, selected view, observed fields,
    target functional predicates, and hidden postcondition outcome for
    a successful one-view \textsc{Grind} episode with a \textsc{Tamp}
    distractor command.
    The trace illustrates the audited decision that the benchmark
    scores: the policy succeeds by selecting the view that exposes the
    command-relevant functional chain before execution.
    This is one representative successful trace; hard $H{=}1$ aggregate
    success is reported in the experiments and supplement.}
    \label{fig:task-conditioned-view-selection}
\end{figure}

The key implementation detail is that $g$ is a deterministic ordinal
score, not a learned RGB scorer or privileged predicate checker.
It uses the command template~$R(\tau)$, the ledger's missing/stale
set~$\mathcal{M}_t$, and public candidate metadata including semantic
view type, camera relation, and nominal functional-part coverage.
This metadata is a strong design aid and is disclosed as part of the
reference method; it does not include hidden state, success labels,
future postconditions, contact logs, or unselected RGB content.
The resulting pipeline---hereafter the \emph{diagnostic reference
selector}---therefore serves as an auditable metadata-assisted anchor
that future learned view selectors can be measured against. \Cref{fig:task-conditioned-view-selection} gives a concrete
logged example of the score--select--execute--verify chain.

\paragraph{Optional action-induced belief revision.}
After a scripted skill executes, the evaluator returns a
postcondition observation.
\aibr{} compares predicted and observed postconditions, localizes the
stale predicate type, lowers confidence in contradicted ledger
entries, and selects a local repair view.
Recovery is charged in the view/step budget and counted only when the
final simulator postcondition passes.
Because successful view selection makes mismatches rare in the
primary protocol, \aibr{} is treated as a repair diagnostic and
stress-test component rather than the main source of the benchmark's
performance gains.
\section{Experimental Setup}

\paragraph{Setup.}
The primary coffee experimental unit is 500~frozen cases (five tasks, 50~scene groups, two variants each), each paired with twelve candidate RGB views (ten in Drawer and Microwave).
We freeze task, postcondition, method, candidate-view, and
failure-code manifests before evaluation.
All non-privileged baselines share the same initial RGB view,
candidate-view list, camera metadata, initial budget, scripted skill
library, simulator reset distribution, and hidden postcondition evaluator.
Hidden simulator predicates, ground-truth postconditions, evaluator
labels, and unselected endpoint renders are never exposed before action.
The supplement specifies each method's sensing contract explicitly.
Applying multiple methods, budgets, and inference repeats yields
60{,}000 closed-loop scripted-skill episodes and 55{,}500
functional-OOD episodes from the primary coffee domain.

All raw records and SHA-256 manifests are retained.

\paragraph{Statistical reporting.}
At the primary budget, each task-method cell contains 300~episodes
(50~scene groups $\times$ 2~variants $\times$ 3~inference repeats);
each aggregate method row contains 1{,}500~episodes across five tasks.
We treat frozen cases---not repeated answers from the same case---as
the conceptual unit for generalization.
The supplement reports paired scene-group bootstrap intervals over
500~case clusters, where each cluster is a task, group, and variant
with three inference repeats averaged before resampling.
Budget-curve bands reproduce source-provided episode-level
normal-approximation 95\% intervals and are descriptive.
The appendix retains the full per-budget, per-task, bootstrap, and
failure-diagnostic plots so that large aggregate gains can be checked
against task-level and case-level behavior.

\paragraph{Baselines.}
Random samples valid views uniformly;
Fixed~scan follows a deterministic camera order;
Pose~coverage chooses geometry-revealing views regardless of command;
History~active reuses previously useful views without an explicit
target ledger;
Target~ledger stores target predicates but does not score views by
missing functional evidence;
Privileged~diagnostic uses privileged view labels under a fixed
diagnostic selector.
This privileged diagnostic is a privileged oracle-map/headroom anchor:
it does not perform adaptive postcondition repair and may be affected
by fixed single-policy tie-breaking.
Success is the percentage of episodes whose scripted skill satisfies
the hidden postcondition.

\paragraph{Diagnostic QA versus actionability.}
VLM question answering verifies whether selected images contain
answerable spatial-functional information.
It is not the main performance claim.
The main evidence is whether selected observations make the downstream
action pass simulator-state postconditions.

\section{Results}

\begin{table*}[t]
\centering
\footnotesize
\setlength{\tabcolsep}{0pt}
\renewcommand{\arraystretch}{0.85}
\begin{tabular}{@{}p{0.19\linewidth}*{9}{>{\raggedleft\arraybackslash}p{0.09\linewidth}}@{}}
\toprule
 & \multicolumn{5}{c}{\cellcolor{tosblue}\textit{Task success (\%)\,$\uparrow$}} & \cellcolor{tosgray} & \multicolumn{3}{c}{\cellcolor{tosorange}\textit{Execution diagnostics}}\\[-2pt]
\cmidrule(lr){2-6}\cmidrule(lr){8-10}
Methods & Tamp & Insert & Place & Serve & Grind & Avg.\,$\uparrow$ & Coll.\,$\downarrow$ & PostFail\,$\downarrow$ & Views\\
\midrule
\multicolumn{10}{c}{\textbf{C.1: Closed-loop actionability}}\\[1pt]
\rowcolor{black!6}\multicolumn{10}{l}{\emph{Passive / generic policies}}\\
\textsc{Random} & 3.7 & 15.3 & 14.7 & 17.0 & 4.7 & 11.1 & 6.2 & 82.7 & 5.0\\
\textsc{Fixed~scan} & 0.0 & 0.0 & 0.0 & 0.0 & 0.0 & 0.0 & 8.1 & 91.9 & 5.0\\
\textsc{Pose~coverage} & 0.0 & 0.0 & 0.0 & 0.0 & 0.0 & 0.0 & 4.4 & 95.6 & 5.0\\
\textsc{History~active} & 0.0 & 40.0 & 28.7 & \textbf{98.3} & 97.7 & 52.9 & 2.3 & 44.7 & 5.0\\
\rowcolor{black!6}\multicolumn{10}{l}{\emph{Belief policies}}\\
\textsc{Target~ledger} & 18.3 & 0.0 & 43.7 & 0.0 & 0.0 & 12.4 & 3.1 & 84.5 & 5.0\\
\textsc{View~Belief~Map} & \textbf{99.3} & \underline{97.0} & \underline{97.7} & \underline{97.3} & \underline{98.3} & \underline{97.9} & \underline{2.1} & \textbf{0.0} & 5.0\\
\textsc{VBM+AIBR} & \underline{98.7} & \textbf{99.3} & \textbf{98.0} & \textbf{98.3} & \textbf{99.7} & \textbf{98.8} & \textbf{1.2} & \textbf{0.0} & 5.0\\
\rowcolor{black!6}\multicolumn{10}{l}{\emph{Privileged diagnostic}}\\
\textsc{Priv.\ diagnostic}$^{\dagger}$ & 98.7 & 99.3 & 97.3 & 97.3 & 97.7 & 98.1 & 1.9 & 0.0 & 5.0\\
\midrule
 & \multicolumn{5}{c}{\cellcolor{tosblue}\textit{Task success (\%)\,$\uparrow$}} & \cellcolor{tosgray} & \multicolumn{3}{c}{\cellcolor{tosorange}\textit{Functional evidence (\%)}}\\[-2pt]
\cmidrule(lr){2-6}\cmidrule(lr){8-10}
Methods & Tamp & Insert & Place & Serve & Grind & Avg.\,$\uparrow$ & Target\,$\uparrow$ & Wrong & Select.\,$\uparrow$\\
\midrule
\multicolumn{10}{c}{\textbf{C.2: Same-object functional OOD}}\\[1pt]
\rowcolor{black!6}\multicolumn{10}{l}{\emph{Category / command-agnostic policies}}\\
\textsc{Random} & 4.0 & 3.7 & 9.3 & 4.0 & 2.7 & 4.7 & 32.9 & 32.6 & 0.3\\
\textsc{Fixed~scan} & 0.0 & 0.0 & 0.0 & 0.0 & 0.0 & 0.0 & 20.0 & 20.0 & 0.0\\
\textsc{Category~shortcut} & 0.0 & 0.0 & 18.3 & 0.0 & 0.0 & 3.7 & 26.7 & 26.7 & 0.0\\
\textsc{Spatial~only} & 0.0 & 0.0 & 0.0 & 0.0 & 0.0 & 0.0 & 16.7 & 23.3 & $-$6.7\\
\textsc{Cmd-agnostic~afford.} & 0.0 & 0.0 & 0.0 & 98.7 & 0.0 & 19.7 & 46.7 & 73.3 & $-$26.7\\
\rowcolor{black!6}\multicolumn{10}{l}{\emph{Command-bound belief policies}}\\
\textsc{View~Belief~Map} & \textbf{99.3} & \textbf{99.7} & \textbf{98.3} & \underline{98.0} & \textbf{98.7} & \underline{98.8} & \textbf{100.0} & 30.0 & \textbf{70.0}\\
\textsc{VBM+AIBR} & \textbf{99.3} & \underline{99.0} & \textbf{98.3} & \textbf{99.0} & \textbf{98.7} & \textbf{98.9} & \textbf{100.0} & 30.0 & \textbf{70.0}\\
\rowcolor{black!6}\multicolumn{10}{l}{\emph{Privileged diagnostic}}\\
\textsc{Priv.\ diagnostic}$^{\dagger}$ & 98.3 & 98.3 & 98.7 & 99.3 & 99.0 & 98.7 & 100.0 & 30.0 & 70.0\\
\bottomrule
\end{tabular}
\caption{\textbf{Main results under both evaluation protocols.}
\emph{C.1} (closed-loop, $H{=}5$): baselines share the same candidate set
and scripted skill interface.
\emph{C.2} (functional OOD, $H{=}3$): the same coffee-machine scene admits
multiple commands; Target and Wrong report functional-cue hits;
Select.\ is their difference.
AIBR occasionally trades a small per-task decrement
(e.g.\ Tamp in C.1) for cross-task repair capability.
\textbf{Bold}: best; \underline{underline}: second-best (non-privileged).
$^{\dagger}$Fixed label-assisted calibration anchor, not an oracle upper bound.}
\label{tab:main-results}
\vspace{-8pt}
\end{table*}

\paragraph{Closed-loop actionability.}
\Cref{tab:main-results} shows that task-conditioned view
selection turns spatial belief into executable action.
The diagnostic reference selector with optional \aibr{} reaches 98.8\%
average success, improving over History~active by 45.9~points and
over Target~ledger by 86.4~points.
The gain is not explained by memory alone:
the ledger must expose which missing predicate should determine the
next view.
The budget curves (\cref{fig:success-budget} in the supplement) confirm that the gap is not an artifact
of choosing a single favorable budget;
the advantage of the diagnostic reference selector persists across
$H\!\in\!\{1,3,5,8\}$ in both the closed-loop and functional-OOD
protocols.


The near-ceiling primary results should not be read as the benchmark
being solved under every observation regime.
The supplement adds a hard $H{=}1$ budget stress split re-summarized
from the frozen episode logs.
At $H{=}1$, recovery-capable methods may still acquire additional
charged observations after mismatch, so the constraint limits the
initial selection, not the total observation cost.
Even with that allowance, the diagnostic reference with \aibr{} drops to 19.0\% in~C.1
and 58.6\% in~C.2, exposing remaining difficulty that is hidden
by the main $H{=}5$ and $H{=}3$ settings.


\paragraph{Same-object functional OOD.}
\Cref{tab:main-results} evaluates the harder same-object setting.
A method may observe the coffee machine and still fail if it observes
the wrong functional part for the command.
The diagnostic reference selector with optional repair reaches
98.9\% success and 70.0 functional selectivity;
category shortcuts and command-agnostic affordance policies fail
despite seeing plausible object parts.
The functional-evidence visualization (\cref{fig:functional-ood} in the supplement) shows the dissociation:
target-functional cue observation tracks success, whereas
wrong-command cue observation does not.
\vspace{-12pt}

\paragraph{Foundation-model stress test.}
We upgrade the foundation-model view-scoring protocol from a
500-row endpoint stress check to a 1{,}500-case strict-clean benchmark.
A VLM is admitted only when all 1{,}500 raw responses have
HTTP~200/endpoint-OK status, strict JSON parses, unique exact-$H{=}3$
selections drawn from the twelve frozen candidates, complete evaluator
records, exact case order, and no fallback parsing.
Seven models pass this gate.
Their postcondition success ranges from 26.0\% to 42.9\%:
Grok~4.3 leads at $42.9\!\pm\!2.5\%$, followed by
Grok~4.6 at $42.5\!\pm\!2.5\%$,
Amazon Nova~Pro at $40.3\!\pm\!2.5\%$,
Qwen3-VL-8B at $39.4\!\pm\!2.5\%$,
Gemma~3 27B at $32.5\!\pm\!2.4\%$,
Llama~4 Maverick at $28.3\!\pm\!2.3\%$,
and Sonar~Pro at $26.0\!\pm\!2.2\%$
(normal-approximation 95\% intervals; $n{=}1{,}500$ per model).
All seven strict-clean VLMs cluster within a 16.9-point band
across multiple provider and model families.
Even the strongest model remains 55.9~points below the diagnostic reference selector
at the primary setting, suggesting a systematic
gap in command-conditioned view selection rather than an idiosyncratic
weakness of one provider.
The gap is not a controlled architecture comparison---the methods
use different priors and sensing contracts---but it shows that current
prompted ranking does not recover the benchmark's command-to-evidence
rules.

The task-level pattern is consistent across models: Tamp and Place, whose target predicates involve large, visually salient parts (portafilter basket, cup--spout alignment), admit near-perfect
scores from several VLMs, whereas Insert, Serve, and Grind require observing smaller or partially occluded functional chains (hopper--chute path, drip-tray clearance, grinder--receiver contact) that prompted VLMs fail to prioritize. This suggests that the gap is driven by the models' inability to bind the command to the specific functional-part evidence chain, not by a blanket failure to parse candidate images.


The stress test comprises 1{,}500 distinct strict-clean
case-level outcomes (300 per task), each evaluated once
under the frozen command-plus-RGB candidate-view contract.
Only models passing the full 1{,}500-case strict-clean gate
enter the reported table; the supplement documents the gate
criteria and lists models whose endpoints were unavailable.

\paragraph{Sanity checks.}
The sanity-checks table (\cref{tab:sanity-checks} in the supplement) consolidates the shortcut and leakage
sanity checks.
These controls are intentionally conservative: they reuse the same
frozen same-object functional-OOD cases rather than introducing
easier auxiliary tasks.
Category shortcuts, command-agnostic affordance viewing, removing
functional fields, and removing target conditioning all fail under
the same hidden evaluator.
This pattern argues against a leakage explanation in which success
comes from object category, hidden postconditions, or a template
rule that ignores the command.

\vspace{-6pt}
\paragraph{Ablation.}
The ablation table (\cref{tab:ablation} in the supplement) confirms that the decisive components are
command binding, functional fields, and target conditioning.
Removing functional fields does not harm the easier C.1 closed-loop
protocol, but it collapses C.2 functional-OOD success to 0.0\%;
the distinction supports the claim that functional evidence matters
specifically when object category is an unreliable shortcut.
Removing \aibr{} has little effect in the primary functional-OOD
split because mismatches are already rare after selecting the correct
functional evidence;
the recovery stress test (\cref{fig:recovery} in the supplement) isolates the setting
where repair matters.
The supplement reports the full task-level ablation matrices.
\paragraph{Three-domain evaluation.}
Under a unified $H{=}3$ protocol across all three domains,
VBM achieves 98.9\% on Coffee (360 cases), 100.0\% on
Drawer/Cabinet (360 cases), and 100.0\% on
Microwave/Appliance (360 cases), yielding 99.6\% overall
(1{,}076/1{,}080; Wilson 95\% CI\@: 99.05--99.86\%).
Random selection yields 5.0\%, 21.4\%, and 15.0\% respectively.
Each domain covers three independent scene seeds with no seed
filtering.
All four residual failures are collision events in the coffee
domain; no domain produces a postcondition failure.
Results use the frozen v2 cross-domain evaluator, which
resolves predicate-definition inconsistencies in an
earlier protocol. The scoring rule and hyperparameters
are shared across domains; domain-specific predicate
schemas, command templates, and public metadata are
instantiated once for each domain before evaluation
and then frozen.
Relative to the preliminary audit, both the evaluator
version and the acquisition budget differ, so their
individual effects are not isolated.

\paragraph{RGB-derived coverage.}
To test whether the reference pipeline depends on simulator-derived
view metadata at inference, we replace its nominal view-to-predicate
coverage mapping with \textsc{VBM-Vision}.
A frozen CLIP-B/32 encoder first converts each candidate RGB image
into a task-agnostic visual signature, and a lightweight per-predicate
linear head estimates functional-part visibility.
The head is trained with visibility labels on training scene groups
and evaluated five-fold out of fold; every reported evaluation image
is scored by a model that did not observe its scene group.
At inference, the visual stage receives neither the command nor
camera/view identifiers, simulator state, success labels, evaluator
outputs, or nominal coverage metadata.
The unchanged command-to-predicate binding is applied only after
visual coverage prediction.
With the evaluator, scripted skills, and postconditions fixed,
\textsc{VBM-Vision} reaches 98.5\% C.1 success at $H{=}5$ and
98.9\% C.2 success at $H{=}3$, with zero postcondition failures
in both protocols.
The remaining 1.5\% and 1.1\% failures are collision events from
the shared scripted skills.
Shuffling the RGB-derived coverage rows within each candidate set
reduces success to 14.9\% and 11.4\%, respectively, confirming that
the result depends on candidate-image content rather than
simulator-derived coverage metadata at inference.

\begin{table}[!t]
\centering\small
\setlength{\tabcolsep}{2.0pt}
\renewcommand{\arraystretch}{1.05}
\begin{tabular}{llrrrrr}
\toprule
Protocol & Coverage source & Succ. & Coll. & PF & Target & Select. \\
\midrule
C.1, $H{=}5$ & VBM-Vision (RGB) & \textbf{98.5} & 1.5 & 0.0 & -- & -- \\
C.1, $H{=}5$ & Shuffled RGB  & 14.9 & 1.2 & 21.1 & -- & -- \\
\midrule
C.2, $H{=}3$ & VBM-Vision (RGB) & \textbf{98.9} & 1.1 & 0.0 & 100.0 & 90.0 \\
C.2, $H{=}3$ & Shuffled RGB & 11.4 & 0.8 & 22.5 & 33.1 & 0.6 \\
\bottomrule
\end{tabular}
\caption{\textbf{Image-derived functional coverage.}
\textsc{VBM-Vision} replaces simulator-derived coverage metadata
at inference with a frozen RGB signature and a scene-group
out-of-fold linear coverage head.
The head is supervised only by predicate-visibility labels from
training folds; the fixed command binding is applied after
task-agnostic visual prediction.
The evaluator, scripted skills, postconditions, candidate sets,
and budgets are unchanged.
\emph{Shuffled RGB} permutes the predicted coverage rows among
the twelve candidates within each case; its collapse verifies
that the positive result depends on candidate-image content.
Succ.\ = success (\%); Coll.\ = collision (\%);
PF = postcondition failure (\%);
Target = target-cue rate (\%);
Select.\ = target-cue rate $-$ wrong-command-cue rate.
Each row aggregates 4{,}500 evaluator episodes.}
\label{tab:vbm-vision}
\vspace{-8pt}
\end{table}

\section{Discussion and Conclusion}

Seeing What Execution Needs reframes spatial reasoning for simulated manipulation as actionability: whether an agent acquires the evidence required by the current command before a shared skill executes. Same-object functional OOD, shortcut controls, and the hard $H{=}1$ initial-selection stress test show that object-level or wrong-command evidence is insufficient. Across their respective sensing contracts, seven strict-clean VLMs remain 55.9 points below the diagnostic reference.

The current benchmark operates within a specific scope: frozen RGB candidate views, shared scripted skills, and simulator-state postconditions. The diagnostic reference uses semantic view names and nominal functional-part coverage and is neither a learned perception model nor an optimal selector; the VLM and reference results accordingly reflect different sensing contracts and do not constitute a controlled architecture comparison. The primary C.1/C.2 track contains 500 frozen coffee cases; the aligned cross-domain track adds 360 cases in each of two additional domains. Across three equal-scale domains (360 cases each), VBM reaches 99.6\% pooled success under frozen domain-specific schemas. Broader evaluation with additional object families, real-robot execution, and fully learned RGB-based view-selection policies remains future work. Future VLM/VLA benchmarks should measure not only what a model knows, but whether it can acquire the evidence the next action requires.

\FloatBarrier


\paragraph{Acknowledgments.}
We thank Guangyu Li (Hong Kong University of Science and Technology / Mohamed bin Zayed University of Artificial Intelligence) for providing the initial simulation draft that supported early development of this benchmark.

{
    \small
    \bibliographystyle{ieeenat_fullname}
    \bibliography{main}
}

\clearpage
\setcounter{page}{1}
\maketitlesupplementary
\appendix

\setcounter{topnumber}{8}
\setcounter{bottomnumber}{5}
\setcounter{totalnumber}{12}
\setcounter{dbltopnumber}{5}
\renewcommand{\topfraction}{0.98}
\renewcommand{\bottomfraction}{0.95}
\renewcommand{\textfraction}{0.02}
\renewcommand{\floatpagefraction}{0.65}
\renewcommand{\dbltopfraction}{0.98}
\renewcommand{\dblfloatpagefraction}{0.65}
\setlength{\floatsep}{8pt plus 2pt minus 1pt}
\setlength{\textfloatsep}{10pt plus 2pt minus 2pt}
\setlength{\intextsep}{4pt plus 2pt minus 1pt}
\setlength{\dblfloatsep}{4pt plus 2pt minus 1pt}
\setlength{\dbltextfloatsep}{6pt plus 2pt minus 2pt}
\makeatletter
\setlength{\@fptop}{0pt}
\setlength{\@fpsep}{5pt}
\setlength{\@fpbot}{0pt plus 1fil}
\setlength{\@dblfptop}{0pt}
\setlength{\@dblfpsep}{5pt}
\setlength{\@dblfpbot}{0pt plus 1fil}
\makeatother

\providecommand{\tabletoprule}{\toprule[0.8pt]}
\providecommand{\tablemidrule}{\midrule[0.4pt]}
\providecommand{\tablebottomrule}{\bottomrule[0.8pt]}
\providecommand{\widefig}[4]{%
\begin{figure*}[!t]
    \centering
    \includegraphics[width=#1\linewidth]{#2}
    \caption{#4}
    \label{#3}
\end{figure*}}
\providecommand{\onefig}[4]{%
\begin{figure}[!tbp]
    \centering
    \includegraphics[width=#1\linewidth]{#2}
    \caption{#4}
    \label{#3}
\end{figure}}

\section{Supplementary Scope}
\label{app:organization}

The main paper carries the central claim: task-conditioned spatial-functional evidence selection improves closed-loop manipulation under a fixed scripted-skill interface. This supplement contains every distinct scientific result and audit boundary supporting the paper: the information boundary, baseline fairness contract, paired scene-group bootstrap, strict-clean seven-model evidence, charged-view costs, aggregate and task-level ablations, VBM-Vision method details, three-domain evaluation, hard one-view stress split, logged rollout trace, raw MuJoCo grounding frames, and compact task/failure diagnostics.

We follow three rules, summarized in Table~\ref{tab:supp-scope}. First, illustrative mechanism figures and empirical evidence are separated. The main paper's hybrid schematics explain the benchmark interface; the supplement's quantitative plots and tables trace to frozen rollouts, CSV summaries, or packaged manifests. Second, strict-clean, primary-run, and ablation-run records remain explicitly separated. Third, a value set is shown once in its most auditable form; complementary plots are retained only when they expose a relationship not legible from an exact table.

\begin{table*}[!t]
\centering
\small
\setlength{\tabcolsep}{4pt}
\renewcommand{\arraystretch}{1.02}
\begin{tabularx}{\textwidth}{>{\raggedright\arraybackslash}p{0.22\textwidth}X}
\tabletoprule
Retained evidence & Why it stays in the supplement \\
\tablemidrule
Raw MuJoCo filmstrips & Ground the benchmark in actual simulator RGB frames, especially because the main figures are hybrid schematics. \\
Baseline fairness and metrics contract & Makes clear which methods see commands, ledgers, functional fields, AIBR, or privileged labels. \\
Sanity-checks and ablation tables & Detailed control, component-removal, and ablation tables supporting the main-paper text paragraphs. \\
Paired scene-group bootstrap & Checks that primary gains do not come from repeated answers on the same case family. \\
Strict-clean VLM protocols & Documents the seven-model 1,500-case strict-clean gate and results (Tables~\ref{tab:strict-vlm-gate}--\ref{tab:strict-vlm-overall} and Table~\ref{tab:strict-vlm-task-main}); earlier 500-row records are archived in the release artifact. \\
Charged-view and ablation evidence & Completes the cost and component analyses promised by the main paper without substituting primary- and ablation-run values. \\
Budget-curve and diagnostic visualizations & Budget-sweep, functional-OOD evidence, and recovery stress-test figures complementing the main results table. \\
Three-domain evaluation & Full cross-domain VBM evaluation at equal scale (360 cases per domain). \\
Hard $H=1$ stress split and logged trace & Shows residual difficulty under tight budgets and gives an auditable single-case score/\allowbreak select/verify trace. \\
Per-task and failure diagnostics & Checks whether aggregate results hide one easy task family or unsafe failure mode. \\
Reproducibility record & SHA-256 manifests, package validation, chart verification, and provenance chain for regeneration. \\
\tablebottomrule
\end{tabularx}
\caption{\textbf{Supplement roadmap.} Every distinct scientific result and audit boundary is retained once; duplicate presentations are removed.}
\label{tab:supp-scope}
\end{table*}

\section{Benchmark Construction and Information Boundary}
\label{app:benchmark-construction}

The benchmark is a task-conditioned analogue of spatial-belief probing. Instead of asking whether a model can build a complete scene-level cognitive map, each episode asks whether a policy acquires the spatial and functional evidence needed by the current operation. The five primary coffee-task families are \textsc{Tamp}, \textsc{Insert}, \textsc{Place}, \textsc{Serve}, and \textsc{Grind}. Each family contains 50 scene groups and two variants, giving 100 variants per task and 500 variants per protocol before method and inference-repeat expansion.

Each episode contains a hidden simulator state, a command, a fixed set of candidate views, and an observation budget. The agent receives the command, acquired RGB frames, view metadata, and its own belief state. It does not receive hidden simulator predicates, evaluator labels, unselected endpoint renders, contact logs, or postcondition truth before acting. The same scripted skill library is used across methods, so differences in closed-loop success come from information selection and belief revision rather than from changing the low-level controller.

\section{Raw MuJoCo Rendering Examples}
\label{app:raw-mujoco-renderings}

\begin{figure*}[!t]
  \centering
  \IfFileExists{img/fig_mujoco_five_tasks_phase_b.pdf}{%
    \includegraphics[width=\linewidth,trim=36bp 18bp 35bp 12bp,clip]{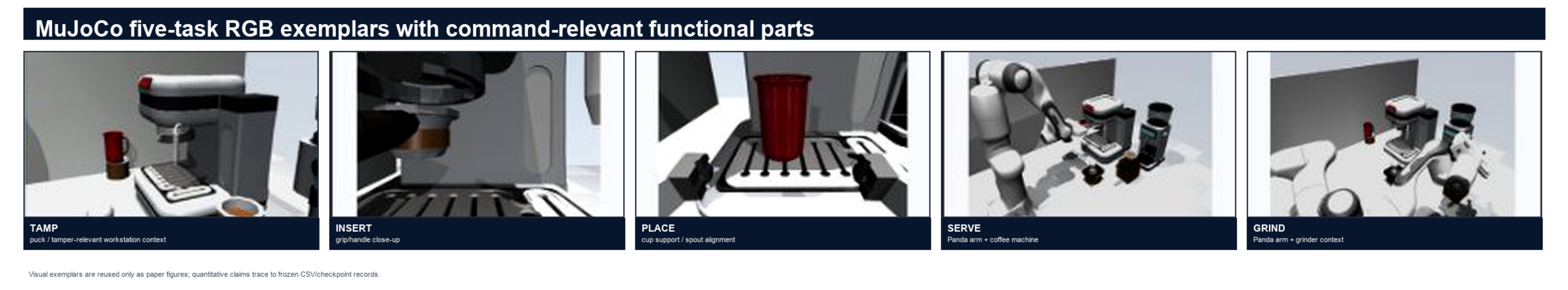}%
  }{%
    \fbox{\parbox{0.92\linewidth}{\centering Five-task MuJoCo/Panda coffee visual-domain figure.}}%
  }
  \caption{\textbf{Primary five-task visual domain.} Raw MuJoCo/Panda RGB exemplars from the coffee workstation show that tamping, insertion, placement, serving, and grinding require different command-relevant functional evidence even when the object family is shared.}
  \label{fig:five-task-rgb}
\end{figure*}

\begin{figure}[!tbp]
  \centering
  \includegraphics[width=\linewidth]{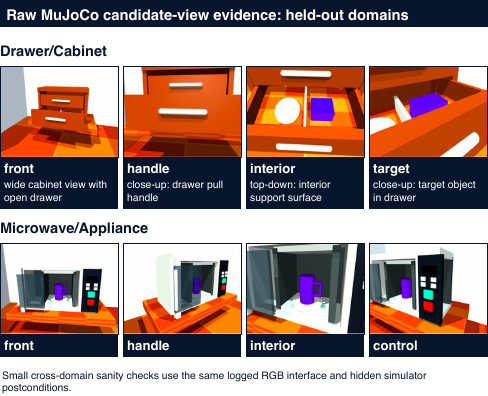}
  \caption{\textbf{Raw MuJoCo candidate-view examples across three domains.} Drawer/Cabinet and Microwave/Appliance use the same logged RGB candidate-view interface and hidden simulator-state postcondition evaluator as the coffee benchmark.}
  \label{fig:app-raw-mujoco-heldout}
\end{figure}

The main-paper figures use hybrid schematics to explain the task-conditioned evidence mechanism. For auditability, Figs.~\ref{fig:five-task-rgb} and~\ref{fig:app-raw-mujoco-heldout} provide raw MuJoCo RGB frames from the frozen simulator interface. These filmstrips are not intended as photorealistic renderings; they document the candidate views, task labels, and held-out domains used by the benchmark. The released artifact additionally includes short qualitative GIF/MP4 demos with raw rollout JSON; quantitative claims in the paper are computed from frozen CSV postconditions, not from these videos.

\section{Protocols, Metrics, and Baseline Fairness}
\label{app:metrics}

\paragraph{Protocols.}
C.1 is the closed-loop actionability protocol. Its primary budget is $H=5$, and it evaluates whether selected observations support successful execution of the shared scripted skill. C.2 is the same-object functional-OOD protocol. Its primary budget is $H=3$, and it evaluates whether methods observe the command-relevant functional cue rather than a plausible cue for a different command. Recovery experiments evaluate postcondition mismatch handling after an initially stale or contradicted belief.

\paragraph{Metrics.}
Task success is the percentage of episodes whose shared scripted skill satisfies the hidden simulator-state postcondition. Collision rate and postcondition-failure rate are reported separately. Average charged views count all acquired views, including recovery observations. Primitive steps count scripted skill steps. Unknown/abstain is tracked separately from failed execution. In C.2, target-functional hit measures whether selected views expose the functional cue required by the current command, wrong-command hit measures whether selected views expose a cue for another command, and command selectivity is their difference in percentage points.

\begin{table*}[!t]
\centering
\small
\setlength{\tabcolsep}{2pt}
\renewcommand{\arraystretch}{1.02}
\begin{tabularx}{\textwidth}{@{}>{\raggedright\arraybackslash}p{0.17\textwidth}>{\centering\arraybackslash}p{0.045\textwidth}>{\centering\arraybackslash}p{0.055\textwidth}>{\centering\arraybackslash}p{0.085\textwidth}>{\centering\arraybackslash}p{0.065\textwidth}>{\centering\arraybackslash}p{0.08\textwidth}X@{}}
\tabletoprule
Method & Cmd. & Ledger & Functional fields & AIBR & Privileged & Purpose \\
\tablemidrule
\textsc{Random} & \ding{55} & \ding{55} & \ding{55} & \ding{55} & \ding{55} & Chance valid-view policy under the same view budget. \\
\textsc{Fixed scan} & \ding{55} & \ding{55} & \ding{55} & \ding{55} & \ding{55} & Deterministic camera-order policy; tests whether coverage alone solves the benchmark. \\
\textsc{Pose coverage} & \ding{55} & \ding{55} & \ding{55} & \ding{55} & \ding{55} & Uses candidate-pose geometry only; no hidden visibility or task-state labels. \\
\textsc{History active} & \ding{51} & \ding{55} & \ding{55} & \ding{55} & \ding{55} & Active view selection from observation history, without structured missing-predicate bookkeeping. \\
\textsc{Target ledger} & \ding{51} & \ding{51} & Partial & \ding{55} & \ding{55} & Stores target predicates but does not score views by missing functional evidence. \\
\tablemidrule
\textsc{View Belief Map} & \ding{51} & \ding{51} & \ding{51} & \ding{55} & \ding{55} & Scores candidate views by expected evidence gain over missing target-relevant predicates. \\
\textsc{VBM+AIBR} & \ding{51} & \ding{51} & \ding{51} & \ding{51} & \ding{55} & Optional local repair after postcondition mismatch; not the main source of primary-split gains. \\
\tablemidrule
\textsc{Priv. diagnostic} & \ding{51} & \ding{51} & \ding{51} & Optional & \ding{51} & Privileged diagnostic selector; not a mathematical upper bound and excluded from non-privileged method ranking. \\
\tablebottomrule
\end{tabularx}
\caption{\textbf{Baseline fairness contract.} All non-privileged methods operate under the same candidate views, budgets, scripted skills, and hidden evaluator. ``Cmd.'' denotes whether the command is available to the selection rule; ``Privileged'' denotes access to hidden state, hidden visibility, or postcondition truth before action.}
\label{tab:supp-baseline-fairness}
\end{table*}

\paragraph{Shortcut controls.}
Table~\ref{tab:supp-baseline-fairness} states the common access contract. Table~\ref{tab:sanity-checks} is a same-object shortcut stress test. Category shortcut, command-agnostic affordance, command-only/no-image, no-functional-fields, and no-target-conditioning controls all use the same hidden postcondition evaluator as the VBM+AIBR. Their failures rule out several common leakage explanations: object category alone, plausible but wrong-command affordance evidence, command templates without visual evidence, pose/category evidence without functional predicates, and target selection without command binding.

\begin{table*}[t]
\centering
\footnotesize
\setlength{\tabcolsep}{4pt}
\renewcommand{\arraystretch}{0.95}
\begin{tabularx}{\textwidth}{@{}p{0.18\textwidth}>{\raggedleft\arraybackslash}p{0.055\textwidth}>{\raggedleft\arraybackslash}p{0.06\textwidth}>{\raggedleft\arraybackslash}p{0.055\textwidth}>{\raggedleft\arraybackslash}p{0.06\textwidth}>{\raggedright\arraybackslash}X@{}}
\toprule
 & \multicolumn{4}{c}{\cellcolor{tosgray}\textit{Functional-OOD metrics (\%)}} & \\[-2pt]
\cmidrule(lr){2-5}
Control & Succ.\,$\uparrow$ & Target\,$\uparrow$ & Wrong & Select.\,$\uparrow$ & What the control rules out\\
\midrule
\rowcolor{black!6}\multicolumn{6}{l}{\emph{Metadata-assisted diagnostic reference}}\\
\textsc{Full~method} & \textbf{98.9} & \textbf{100.0} & 30.0 & \textbf{70.0} & The metadata-assisted diagnostic reference succeeds while selecting command-relevant functional evidence much more often than wrong-command evidence.\\[3pt]
\rowcolor{black!6}\multicolumn{6}{l}{\emph{Shortcut and leakage controls}}\\
\textsc{Category~shortcut} & 3.7 & 26.7 & 26.7 & 0.0 & Seeing the correct object category is insufficient when the same object supports multiple commands.\\[3pt]
\textsc{Cmd-agnostic~afford.} & 19.7 & 46.7 & 73.3 & $-$26.7 & Generic affordance viewing can expose plausible but wrong-command parts and still fail the postcondition.\\[3pt]
\textsc{Command-only/no-image} & 0.0 & 0.0 & 0.0 & 0.0 & Removing all RGB evidence eliminates visual verification, ruling out command-template execution.\\[3pt]
\rowcolor{black!6}\multicolumn{6}{l}{\emph{Component removal}}\\
w/o functional fields & 0.0 & 16.7 & 23.3 & $-$6.7 & The score cannot be explained by pose-only or object-only evidence.\\[3pt]
w/o target conditioning & 3.5 & 26.7 & 26.7 & 0.0 & Removing command-conditioned missing-predicate selection destroys the useful-view advantage.\\
\bottomrule
\end{tabularx}
\caption{\textbf{Sanity checks for information leakage and shortcut behavior.}
All rows use the same-object functional-OOD hidden evaluator and
1,500 episodes per method.
Target cue and Wrong cue are separately measured functional-cue hits;
Select.\ is their difference.
The command-only/no-image row reruns the primary cases with all RGB
evidence removed.
The controls verify that the main success is not explained by category
recognition, command-agnostic affordance viewing, pose-only evidence,
command templates, or access to hidden postconditions.}
\label{tab:sanity-checks}
\end{table*}

\section{Paired Scene-Group Bootstrap}
\label{app:bootstrap}

Table~\ref{tab:scene-group-bootstrap} reports paired scene-group bootstrap intervals using the regenerated episode-level records. For each frozen case, identified by task, group, and variant, we average the three inference repeats for each method. We then resample the 500 paired case clusters with replacement for 10,000 bootstrap replicates. This directly addresses whether the primary aggregate gains depend on repeated answers from the same case. The main conclusion is stable: the View Belief Map with optional AIBR strongly improves over History active in C.1 and over command-agnostic affordance in C.2. The optional AIBR versus View Belief Map rows are small, as expected; they show that the functional-OOD gain is driven mainly by command-conditioned view selection rather than by postcondition repair.

\begin{table*}[!t]
\centering
\small
\setlength{\tabcolsep}{4.0pt}
\renewcommand{\arraystretch}{1.02}
\begin{tabularx}{\textwidth}{llrrX}
\tabletoprule
Protocol & Paired comparison & Cases & Mean diff. & 95\% CI / interpretation \\
\tablemidrule
C.1 & \textsc{VBM+AIBR} -- \textsc{History active} & 500 & 45.9 & [41.9, 49.8], large gain. \\
C.1 & \textsc{VBM+AIBR} -- \textsc{View Belief Map} & 500 & 0.9 & [-0.1, 1.8], small AIBR increment. \\
\tablemidrule
C.2 & \textsc{VBM+AIBR} -- \textsc{Cmd-agnostic afford.} & 500 & 79.1 & [75.7, 82.6], large gain. \\
C.2 & \textsc{VBM+AIBR} -- \textsc{View Belief Map} & 500 & 0.1 & [-0.7, 0.8], small AIBR increment. \\
C.2 & \textsc{VBM+AIBR} -- \textsc{Command only / no image} & 500 & 98.9 & [98.3, 99.4], rules out command-only execution. \\
\tablebottomrule
\end{tabularx}
\caption{\textbf{Paired scene-group bootstrap over frozen cases.} Differences are success-rate percentage points. We average the three inference repeats for each frozen case, then resample the 500 paired case clusters (task, group, variant) with replacement for 10,000 bootstrap replicates.}
\label{tab:scene-group-bootstrap}
\end{table*} 

\newpage

\section{Foundation-Model Stress Test and Three-Domain Evaluation}
\label{app:review-checks}

\subsection{Strict-clean seven-model view-scoring evidence}
\label{app:strict-vlm}

Each case presents the command and all twelve frozen RGB candidates, then asks the model to return an ordered set of exactly three unique view IDs. This is a full-preview ranking diagnostic, not an equal-information comparison with sequential acquisition. A promoted row must pass endpoint/HTTP status, strict JSON, exact unique $H=3$, candidate membership, complete evaluator record, exact case order, and no-fallback checks for all 1,500 cases. The unchanged hidden simulator-state evaluator scores the selected views. Earlier 500-row endpoint records are archived in the release artifact; the strict-clean protocol supersedes them. Exact InternVL3-38B and Molmo-72B endpoints were recorded as unavailable under the same endpoint protocol; no numerical rows are invented or substituted for them.

\begin{table*}[!t]
\centering\scriptsize
\renewcommand{\arraystretch}{1.05}
\begin{tabularx}{\textwidth}{p{0.12\textwidth}p{0.25\textwidth}Xp{0.12\textwidth}}
\tabletoprule
Gate stage & Requirement & Retained evidence & Promoted condition \\
\tablemidrule
Dry-run probe & Five frozen probes return strict JSON with exactly three unique candidate IDs. & Provider smoke/probe logs; a failed endpoint is not promoted. & 5/5 \\
Strict prompt & Exactly three unique IDs, all among the twelve frozen candidates. & Prompt template and model-specific run manifest. & 1,500/1,500 \\
Frozen manifest & Same order, 300 cases/task, twelve RGB candidates/case. & Strict manifest and render audit; 18,000 hashed RGB images. & 1,500/1,500 \\
Endpoint retry & Retry only failed endpoint rows; endpoint and format repair stay separate. & Retry logs and manifests retained without overwriting originals. & Clean rows only \\
Raw responses & One retained raw response per promoted row. & \texttt{raw\_responses.jsonl}; failed/fallback rows remain separate. & 1,500/1,500 \\
Parsed selections & Strict JSON only; no regex/fallback extraction. & Parsed selections and strict-gate JSON. & 1,500/1,500 \\
Episode results & Unchanged hidden postcondition evaluator. & Episode results plus task/overall summaries. & 1,500/1,500 \\
Combined provenance & Combine only clean original/retry/repair rows, preserving order. & Provenance audit retains the original case order. & Exact order \\
SHA256/checks & Hash-check manifests and RGB records. & Manifest/render audits and package SHA256 records. & Pass \\
\tablebottomrule
\end{tabularx}
\caption{\textbf{Strict-clean VLM promotion checklist.} This is the archived admission contract, not a claim that the present document rebuild reran remote model responses.}
\label{tab:strict-vlm-gate}
\end{table*}

\begin{table}[!t]
\centering\scriptsize
\setlength{\tabcolsep}{3.0pt}
\resizebox{\columnwidth}{!}{%
\begin{tabular}{lrrrrrrr}
\tabletoprule
Model & Success $\uparrow$ & Target $\uparrow$ & Wrong $\downarrow$ & Select. $\uparrow$ & Coll. $\downarrow$ & Post-fail $\downarrow$ & $n$ \\
\tablemidrule
Gemma 3 27B & $32.5\pm2.4$ & 63.8 & 28.1 & 35.6 & 3.7 & 63.7 & 1,500 \\
Grok 4.3 & $42.9\pm2.5$ & \textbf{85.2} & 28.2 & \textbf{57.0} & 2.6 & \textbf{54.5} & 1,500 \\
Grok 4.6 & $42.5\pm2.5$ & 80.2 & 27.8 & 52.4 & \textbf{2.4} & 55.1 & 1,500 \\
Llama 4 Maverick & $28.3\pm2.3$ & 66.3 & 32.3 & 34.0 & 3.1 & 68.6 & 1,500 \\
Amazon Nova Pro & $40.3\pm2.5$ & 57.0 & \textbf{20.3} & 36.7 & 3.5 & 56.1 & 1,500 \\
Qwen3-VL-8B & $39.4\pm2.5$ & 66.7 & 20.7 & 45.9 & 3.2 & 57.4 & 1,500 \\
Sonar Pro & $26.0\pm2.2$ & 54.3 & 34.4 & 19.9 & 3.5 & 70.5 & 1,500 \\
\tablebottomrule
\end{tabular}%
}
\caption{\textbf{Verified strict-clean foundation-model view scoring.} Values are percentages except $n$; $\pm$ is the archived normal-approximation 95\% interval half-width. Every row averages 3.0 selected views and 2.6 primitive steps. Target/wrong cue and selectivity are evaluator diagnostics; success is the hidden postcondition outcome. Diagnostics were rounded independently.}
\label{tab:strict-vlm-overall}
\end{table}

\begin{table}[!t]
\centering\scriptsize
\setlength{\tabcolsep}{2pt}
\resizebox{\columnwidth}{!}{%
\begin{tabular}{lrrrrrr}
\toprule
Strict-clean VLM & Tamp & Insert & Place & Serve & Grind & Overall \\
\midrule
Gemma 3 27B & 63.3 & 0.0 & 99.3 & 0.0 & 0.0 & $32.5\!\pm\!2.4$ \\
Grok 4.3 & 67.0 & 10.7 & 99.3 & 37.3 & 0.0 & $42.9\!\pm\!2.5$ \\
Grok 4.6 & 97.7 & 0.0 & 97.3 & 3.7 & 13.7 & $42.5\!\pm\!2.5$ \\
Llama 4 Maverick & 97.3 & 0.0 & 43.7 & 0.3 & 0.0 & $28.3\!\pm\!2.3$ \\
Amazon Nova Pro & 97.7 & 0.0 & 99.7 & 4.3 & 0.0 & $40.3\!\pm\!2.5$ \\
Qwen3-VL-8B & 97.7 & 0.0 & 99.3 & 0.0 & 0.0 & $39.4\!\pm\!2.5$ \\
Sonar Pro & 98.0 & 0.0 & 14.3 & 16.7 & 1.0 & $26.0\!\pm\!2.2$ \\
\bottomrule
\end{tabular}%
}
\caption{\textbf{Strict-clean VLM postcondition success by task.}
Each model is evaluated on $1{,}500$ cases (300 per task) under the same
frozen candidate-view and hidden-postcondition contract.
Overall includes normal-approximation 95\% intervals.
Task-level asymmetry is pronounced: models that approach perfect
success on Tamp or Place can remain near zero on Insert, Serve,
or Grind.
Admission gate and complete diagnostics appear in
\Cref{tab:strict-vlm-gate,tab:strict-vlm-overall}.}
\label{tab:strict-vlm-task-main}
\end{table}

\begin{figure*}[!t]
\centering
\includegraphics[width=0.96\linewidth]{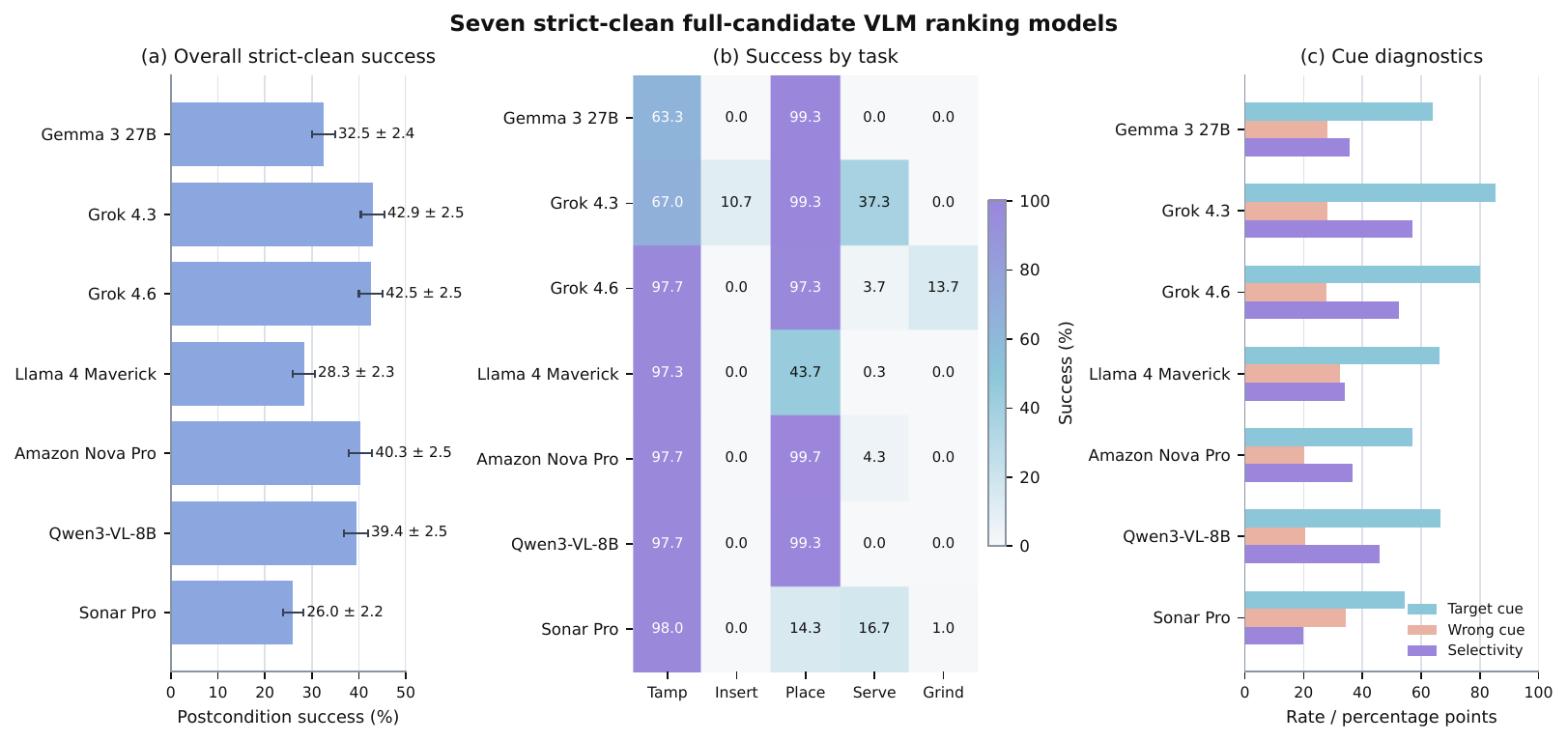}
\caption{\textbf{Seven-model strict-clean comparison.} (a) Overall postcondition success with archived normal-approximation 95\% half-width. (b) Per-task success. (c) Target-cue, wrong-cue, and selectivity diagnostics. Generated only from the archived strict-clean CSVs; it does not equate the VLM full-preview contract with sequential acquisition.}
\label{fig:supp-vlm-comparison}
\end{figure*}
Per-task strict-clean results are reported in Table~\ref{tab:strict-vlm-task-main}. Tables~\ref{tab:strict-vlm-gate}--\ref{tab:strict-vlm-overall} retain the admission contract and full cue diagnostics.

\paragraph{VBM-Vision method details.}
We compute a frozen CLIP-B/32 image-text signature for each of the 18{,}000 candidate RGB records and fit one class-balanced
logistic-regression head per functional predicate. We use five scene-group folds (150 groups total; 120 training and 30 held-out groups per fold), fixed $C{=}1.0$, the \texttt{liblinear}
solver, and no evaluator-success-based hyperparameter tuning.
All reported coverage scores are strictly out of fold. The expanded VBM-Vision archive contains 1{,}500 cases
from 150 scene groups and 18{,}000 candidate RGB images;
all 18{,}000 files have distinct SHA-256 hashes.
Evaluation over three inference seeds yields 4{,}500
episodes per protocol--variant row. All cases from the
same scene group remain in the same fold. The coverage head is trained using nominal predicate-visibility labels from the training folds, but at inference it receives only frozen RGB-derived features and candidate-relative normalizations. It does not receive task or command identity, view or camera identifiers, simulator state, success labels, evaluator outputs, or the nominal view-to-predicate coverage mapping. The command is bound to predicted predicates only after this task-agnostic visual stage.

\paragraph{Three-domain evaluation.}
To assess portability of the benchmark interface beyond the
coffee workstation, we evaluate VBM on two additional MuJoCo
domains: Drawer/Cabinet and Microwave/Appliance, each with 360
cases. Coffee uses 12 candidate views per case (4{,}320 images);
Drawer and Microwave use 10 (3{,}600 each), yielding 11{,}520
candidate images in total. Table~\ref{tab:p2-domain-protocol}
summarizes the task types and hidden postcondition checks.
The selector architecture and hidden postcondition interface
are reused across domains. The scoring rule and hyperparameters
are shared, while domain-specific predicate schemas, command
templates, and public metadata are instantiated once per domain
before evaluation and then frozen.

Under a unified $H{=}3$ budget, VBM achieves
98.9\% on Coffee, 100.0\% on Drawer/Cabinet, and 100.0\% on
Microwave/Appliance, yielding 99.6\% overall
(1{,}076/1{,}080; Wilson 95\% CI: 99.05--99.86\%). Random selection yields
5.0\%, 21.4\%, and 15.0\% respectively. All four residual
failures are collision events in the coffee domain; no domain
produces a postcondition failure. Each domain covers three
independent scene seeds with no seed filtering.
Drawer/Cabinet covers opening, retrieval, and placement;
Microwave/Appliance covers opening, placement, and activation.

\paragraph{Protocol revision.}
The original cross-domain evaluator (v1) conflated pre-action
evidence predicates with post-action postconditions and misapplied
collision risk to non-contact commands. The revised protocol (v2)
corrects these predicate definitions without changing the frozen
RGB images, case identifiers, scene seeds, or success criteria.
Coffee episodes retain the upstream source identifier
\texttt{mujoco\_protocol\_v1}; the protocol-v2 correction applies
only to cross-domain evidence/outcome alignment. Original v1
results are retained in the release package for audit.

\begin{table*}[!t]
\centering
\small
\setlength{\tabcolsep}{5pt}
\renewcommand{\arraystretch}{1.02}
\begin{tabularx}{\textwidth}{p{0.18\textwidth}p{0.24\textwidth}X}
\tabletoprule
Domain & Task types & Hidden postcondition examples \\
\tablemidrule
Drawer/Cabinet & open drawer; retrieve object; place object & drawer openness; handle access; target visibility/reachability after opening; placement on commanded support within distance threshold. \\
Microwave/Appliance & open door; place item; activate control & door open/closed state; item inside or aligned with cavity/support; commanded control/button state; collision and wrong-control failures tracked separately. \\
\tablebottomrule
\end{tabularx}
\caption{\textbf{Held-out domain protocol details.} Each domain contains 360 cases under unified $H{=}3$. Coffee uses 12 candidate views per case (4{,}320 images); Drawer and Microwave use 10 (3{,}600 each); 11{,}520 candidate images in total. Drawer/Cabinet and Microwave/Appliance use articulated-object states and postcondition checks that differ from the coffee workstation.}
\label{tab:p2-domain-protocol}
\end{table*}

\begin{table}[!htbp]
\centering
\small
\setlength{\tabcolsep}{3.0pt}
\renewcommand{\arraystretch}{1.02}
\begin{tabular*}{\columnwidth}{@{\extracolsep{\fill}}lrrrr@{}}
\tabletoprule
Method & Coffee & Drawer & Microwave & Avg. \\
\tablemidrule
Random view & 5.0 & 21.4 & 15.0 & 13.8 \\
\tablemidrule
\textsc{VBM} & 98.9 & 100.0 & 100.0 & 99.6 \\
\tablemidrule
\textsc{Priv. diagnostic} & 98.3 & 100.0 & 100.0 & 99.4 \\
\tablebottomrule
\end{tabular*}
\captionof{table}{\textbf{Three-domain evaluation under unified $H{=}3$ budget.}
Each domain contains 360 cases. This track evaluates VBM portability. VBM+AIBR is omitted because no repairable mismatch is triggered at $H{=}3$; AIBR is evaluated separately under constructed mismatch (\cref{fig:recovery}). Wilson 95\% CI for pooled VBM: 99.05--99.86\%.}
\label{tab:p2-cross-domain}
\end{table}

\vspace{4pt}

\section{Hard Budget and Logged-Trace Evidence}
\label{app:hard-trace}

\begin{figure}[!t]
    \centering
    \includegraphics[width=\linewidth,trim=3bp 5bp 2bp 3bp,clip]{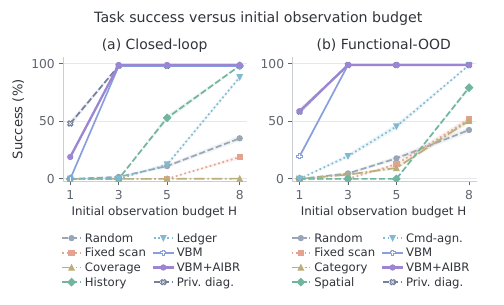}
    \caption{\textbf{Task success versus initial observation budget.}
    Lines connect only the observed budgets $H\!\in\!\{1,3,5,8\}$.
    Recovery methods may use extra charged recovery observations
    after the initial budget, which are analyzed separately
    in the supplementary charged-view curves.}
    \label{fig:success-budget}
\end{figure}

\begin{figure*}[!t]
\begin{minipage}[t]{\columnwidth}
\vspace{0pt}
\centering
\includegraphics[width=0.95\linewidth]{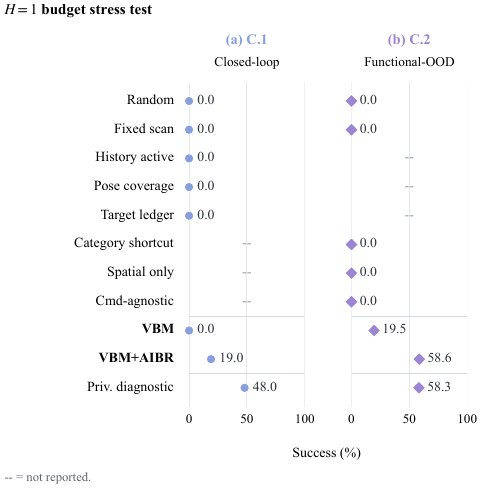}
\caption{\textbf{Hard one-view budget stress test.}
The figure is generated from the frozen episode logs by filtering to $H=1$. It exposes residual difficulty that is obscured by the primary $H=5$ C.1 and $H=3$ C.2 settings. Missing protocol/method combinations are left unconnected rather than imputed. Uncertainty for reported rows is provided in Table~\ref{tab:hard-h1-budget}.}
\label{fig:supp-hard-h1-budget}

\vspace{4pt}
\small
\begingroup
\newcommand{\stat}[2]{\makebox[19pt][r]{#1}\makebox[12pt][c]{$\pm$}\makebox[15pt][l]{#2}}
\begingroup
\fontsize{9}{11}\selectfont
\setlength{\tabcolsep}{0pt}
\renewcommand{\arraystretch}{1.02}
\begin{tabular}{@{}p{0.36\linewidth}>{\centering\arraybackslash}p{0.32\linewidth}>{\centering\arraybackslash}p{0.32\linewidth}@{}}
\tabletoprule
Method & C.1 hard H=1 & C.2 hard H=1 \\
\tablemidrule
Random & \stat{0.0}{0.0} & \stat{0.0}{0.0} \\
Fixed scan & \stat{0.0}{0.0} & \stat{0.0}{0.0} \\
History active & \stat{0.0}{0.0} & -- \\
Pose coverage & \stat{0.0}{0.0} & -- \\
Target ledger & \stat{0.0}{0.0} & -- \\
Category shortcut & -- & \stat{0.0}{0.0} \\
Spatial only & -- & \stat{0.0}{0.0} \\
Cmd-agnostic & -- & \stat{0.0}{0.0} \\
\tablemidrule
VBM & \stat{0.0}{0.0} & \stat{19.5}{2.0} \\
VBM+AIBR & \stat{19.0}{2.0} & \stat{58.6}{2.5} \\
\tablemidrule
Priv. diagnostic & \stat{48.0}{2.5} & \stat{58.3}{2.5} \\
\tablebottomrule
\end{tabular}
\endgroup

\endgroup

\captionof{table}{\textbf{Hard budget stress test.} Success rates at an initial one-view budget, re-summarized from the frozen episode logs. This split exposes saturation hidden by the main H=3/H=5 settings and should be read as a stress diagnostic rather than a replacement for the main protocol. Here $n=1{,}500$ per cell; $\pm$ is the half-width of a normal-approximation 95\% binomial CI; and an en dash denotes a combination absent from the frozen logs. At observed zero, the approximation yields $0.0\pm0.0$ and does not imply zero population uncertainty.}
\label{tab:hard-h1-budget}
\end{minipage}\hfill
\begin{minipage}[t]{\columnwidth}
\vspace{0pt}
\centering
\includegraphics[width=\linewidth]{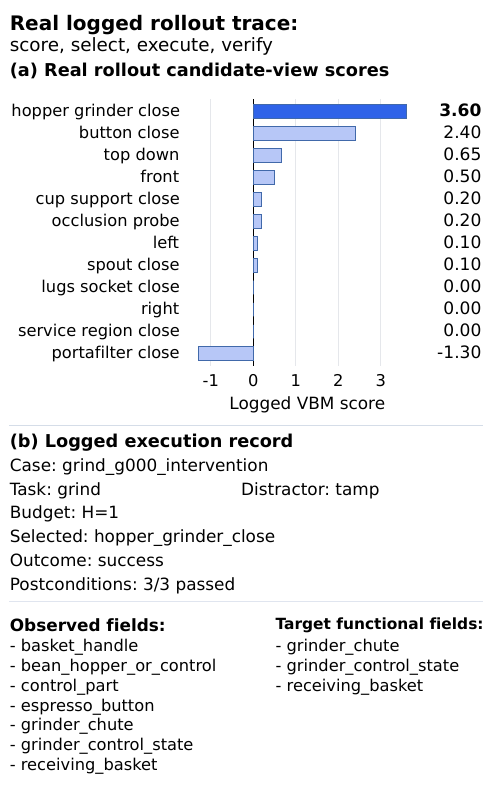}
\caption{\textbf{Real logged rollout trace.} Candidate-view scores, selected view, observed fields, and postcondition outcome are read from the frozen trace record rather than drawn as a storyboard. This is one representative successful trace; the hard $H=1$ aggregate success rates are reported in Table~\ref{tab:hard-h1-budget}. The panel uses the logged score and postcondition records; RGB thumbnails from the frozen image directory are included in the release artifact.}
\label{fig:supp-real-rollout-trace}
\end{minipage}
\end{figure*}

\paragraph{Hard one-view stress split.}
The main protocols use $H=5$ for C.1 and $H=3$ for C.2, where the transparent reference selector approaches saturation. Table~\ref{tab:hard-h1-budget} and Fig.~\ref{fig:supp-hard-h1-budget} expose a harder diagnostic by filtering the frozen episode logs to $H=1$. This is not a new simulator run or a manually edited table; it is a re-summarization of the logged C.1 and C.2 budget-sweep episodes. The one-view constraint makes the task substantially harder: \textsc{VBM+AIBR} reaches 19.0\% in C.1 and 58.6\% in C.2, far below the primary near-ceiling results. Under $H=1$, the privileged diagnostic reaches 48.0\% in C.1 while \textsc{VBM+AIBR} reaches 19.0\%, indicating a substantial headroom for future learned view selectors within the same budget and evaluator.

\paragraph{Logged rollout trace.}
Figure~\ref{fig:supp-real-rollout-trace} replaces a purely conceptual selection illustration with a logged functional-OOD trace. The selected case is a successful \textsc{Grind} intervention episode with a \textsc{Tamp} distractor under $H=1$. The panel shows the actual candidate-view scores stored in the JSONL trace, the selected view, observed fields, target functional predicates, and postcondition pass count. It is therefore a compact audit trail for the central mechanism: the selector succeeds by choosing the view that exposes the command-relevant functional chain.

\section{Charged-View Cost and Ablation Detail}
\label{app:cost-ablation}

\begin{figure*}[!t]
\centering
\includegraphics[width=0.49\linewidth]{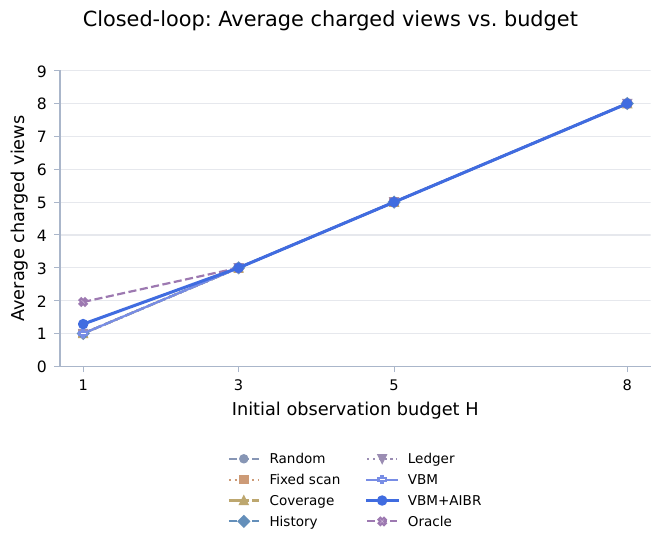}\hfill
\includegraphics[width=0.49\linewidth]{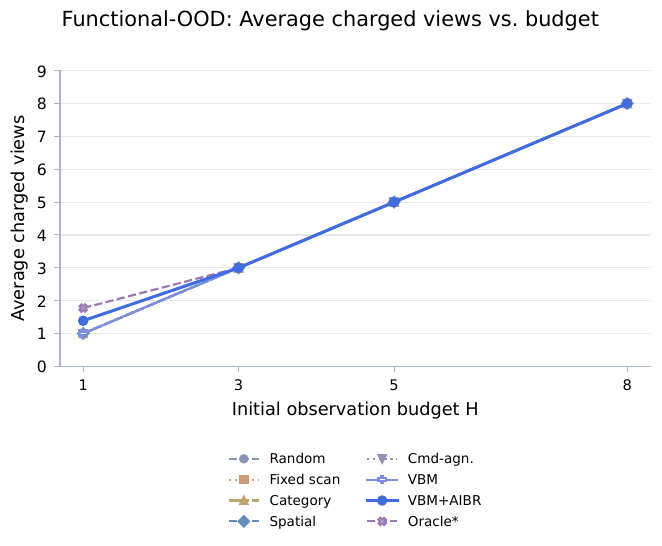}
\caption{\textbf{Average charged views versus initial observation budget.} Left: C.1 closed-loop actionability. Right: C.2 functional-OOD. Recovery observations are charged in addition to the initial selection budget; lines connect only observed budgets $H\in\{1,3,5,8\}$.}
\label{fig:supp-charged-views}
\end{figure*}

\paragraph{Actual charged views across the budget sweep.}
Most methods use exactly the initial budget. Recovery-capable and privileged diagnostic rows can use additional charged observations after mismatch at $H=1$; the full budget curves are shown in Fig.~\ref{fig:supp-charged-views}, and the nontrivial reference rows are listed exactly in Table~\ref{tab:supp-charged-views}. These costs are distinct from success rates.

\begin{table}[!tbp]
\centering\scriptsize
\setlength{\tabcolsep}{3pt}
\begin{tabular*}{\columnwidth}{@{\extracolsep{\fill}}llrrrr@{}}
\tabletoprule
Protocol & Method & $H=1$ & $H=3$ & $H=5$ & $H=8$ \\
\tablemidrule
C.1 & VBM & 1.000 & 3.000 & 5.000 & 8.000 \\
C.1 & VBM+AIBR & 1.287 & 3.000 & 5.000 & 8.000 \\
C.1 & Priv. diagnostic & 1.960 & 3.000 & 5.000 & 8.000 \\
C.2 & VBM & 1.000 & 3.000 & 5.000 & 8.000 \\
C.2 & VBM+AIBR & 1.389 & 3.000 & 5.000 & 8.000 \\
C.2 & Priv. diagnostic & 1.775 & 3.000 & 5.000 & 8.000 \\
\tablebottomrule
\end{tabular*}
\caption{\textbf{Charged-view values for recovery-capable/reference rows.} Other rows in the frozen C.1/C.2 budget summaries use exactly $H$ charged views.}
\label{tab:supp-charged-views}
\end{table}

\paragraph{Ablation-run values.}
Table~\ref{tab:ablation} reports the primary functional-OOD ablation. Table~\ref{tab:supp-c1-ablation} reports the frozen C.1 ablation aggregate. Its full-method value (97.60\%) belongs to the ablation run and must not be substituted for the 98.8\% primary-run value. Table~\ref{tab:supp-task-ablation} exposes the corresponding task structure for C.1 and C.2.

\begin{table}[t]
\centering
\small
\setlength{\tabcolsep}{3.5pt}
\renewcommand{\arraystretch}{1.05}
\resizebox{\linewidth}{!}{%
\begin{tabular}{lrrrrrr}
\toprule
Variant & Succ.\,$\uparrow$ & Target\,$\uparrow$ & Wrong & Select.\,$\uparrow$ & Coll.\,$\downarrow$ & PostFail\,$\downarrow$\\
\midrule
\textbf{VBM+AIBR} & \textbf{98.9} & \textbf{100.0} & 30.0 & \textbf{70.0} & \textbf{1.1} & \textbf{0.0}\\
w/o AIBR & \underline{98.9} & \underline{100.0} & 30.0 & \underline{70.0} & \underline{1.1} & \textbf{0.0}\\
w/o command binding & 19.7 & 46.7 & 73.3 & $-$26.7 & 3.9 & 76.3\\
w/o functional fields & 0.0 & 16.7 & 23.3 & $-$6.7 & 3.5 & 96.5\\
w/o target conditioning & 3.5 & 26.7 & 26.7 & 0.0 & 5.7 & 90.7\\
\bottomrule
\end{tabular}%
}
\caption{\textbf{Ablation under functional-OOD.}
Command binding and functional fields are the decisive components;
without them, success collapses even when the same object category is visible.
The 16.7 value for the no-functional variant is target-cue observation,
not task success.
\textbf{Bold}: best; \underline{underline}: second-best.}
\label{tab:ablation}
\end{table}

\begin{table}[!tbp]
\centering\scriptsize
\setlength{\tabcolsep}{3pt}
\begin{tabular*}{\columnwidth}{@{\extracolsep{\fill}}lrrrrr@{}}
\tabletoprule
C.1 variant & Success & Collision & Post-fail & Views & $n$ \\
\tablemidrule
VBM+AIBR & 97.60 & 2.40 & 0.00 & 5.0 & 1,500 \\
w/o AIBR & 98.13 & 1.87 & 0.00 & 5.0 & 1,500 \\
w/o belief memory & 47.33 & 2.53 & 50.13 & 5.0 & 1,500 \\
w/o functional fields & 99.00 & 1.00 & 0.00 & 5.0 & 1,500 \\
w/o target conditioning & 0.00 & 5.00 & 95.00 & 5.0 & 1,500 \\
\tablebottomrule
\end{tabular*}
\caption{\textbf{Frozen C.1 ablation aggregate.} Values are percentages except views and $n$. This ablation-run summary is kept separate from the primary result.}
\label{tab:supp-c1-ablation}
\end{table}

\begin{table*}[!t]
\centering\small
\setlength{\tabcolsep}{5pt}
\begin{tabular*}{\textwidth}{@{\extracolsep{\fill}}llrrrrrr@{}}
\tabletoprule
Protocol & Variant & Tamp & Insert & Place & Serve & Grind & Average \\
\tablemidrule
C.1 & Full & 97.0 & 98.3 & 96.7 & 98.0 & 98.0 & 97.6 \\
C.1 & w/o AIBR & 96.7 & 98.7 & 98.3 & 98.3 & 98.7 & 98.1 \\
C.1 & w/o belief memory & 0.0 & 39.3 & 2.0 & 97.3 & 98.0 & 47.3 \\
C.1 & w/o functional fields & 99.3 & 99.3 & 98.7 & 98.7 & 99.0 & 99.0 \\
C.1 & w/o target conditioning & 0.0 & 0.0 & 0.0 & 0.0 & 0.0 & 0.0 \\
\tablemidrule
C.2 & Full & 99.0 & 99.0 & 99.0 & 98.7 & 99.0 & 98.9 \\
C.2 & w/o AIBR & 98.3 & 98.7 & 99.0 & 98.3 & 100.0 & 98.9 \\
C.2 & w/o command binding & 0.0 & 0.0 & 0.0 & 98.7 & 0.0 & 19.7 \\
C.2 & w/o functional fields & 0.0 & 0.0 & 0.0 & 0.0 & 0.0 & 0.0 \\
C.2 & w/o target conditioning & 0.0 & 0.0 & 17.7 & 0.0 & 0.0 & 3.5 \\
\tablebottomrule
\end{tabular*}
\caption{\textbf{Task-level ablation success (\%).} Reconstructed directly from the frozen task CSVs; the matrix shows whether an aggregate difference is concentrated in one task.}
\label{tab:supp-task-ablation}
\end{table*}

\section{Compact Diagnostic Figures}
\label{app:compact-diagnostics}

The following figures test whether the main aggregate hides task-family or failure-mode artifacts. They complement, rather than repeat, the exact cost and ablation tables above.

\begin{figure}[!t]
    \centering
    \includegraphics[width=\linewidth,trim=3bp 5bp 2bp 3bp,clip]{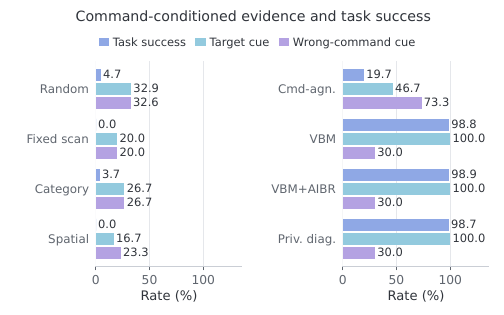}
    \caption{\textbf{Functional evidence must be command-bound.}
    Target-functional cue observation tracks task success,
    whereas wrong-command cue observation does not.
    Target and wrong-command cue rates are separate measurements
    and need not sum to 100\%.}
    \label{fig:functional-ood}
\end{figure}

\begin{figure}[!t]
    \centering
    \includegraphics[width=\linewidth,trim=3bp 3bp 2bp 3bp,clip]{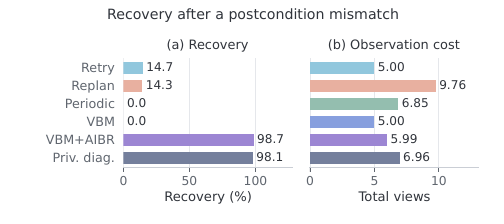}
    \caption{\textbf{Recovery stress test for postcondition mismatch.}
    Optional \aibr{} repairs stale belief fields locally
    and restores 98.7\% of mismatch episodes in the constructed
    mismatch setting.
    The right panel reports total charged views including initial
    observations, not only extra recovery views.}
    \label{fig:recovery}
\end{figure}

\begin{figure*}[!t]
\centering
\includegraphics[width=0.931\linewidth]{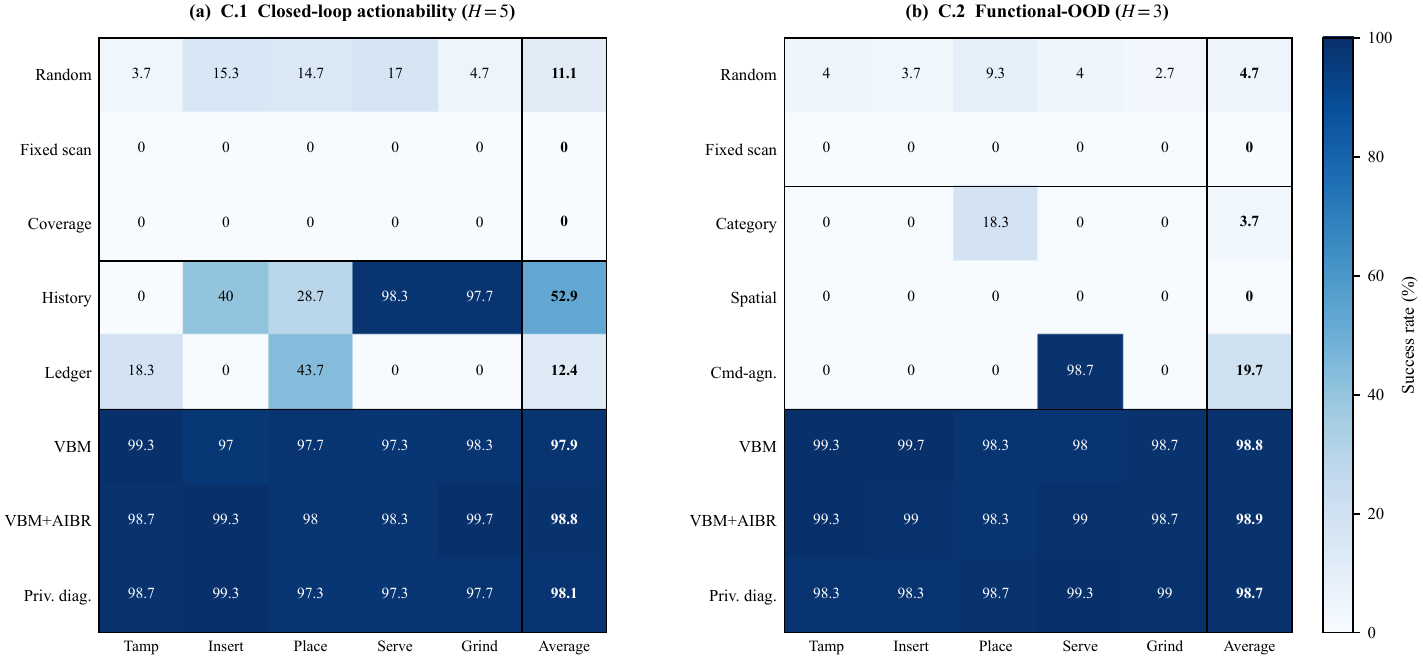}
\caption{\textbf{Closed-loop task success by task at $H=5$ and functional-OOD task success by task at $H=3$.} (a,b) Task-level matrices show the five task families and the corresponding aggregate. Same-object command shifts expose category shortcuts.}
\label{fig:supp-task-combined}
\label{fig:supp-c1-task}
\label{fig:supp-c2-task}
\end{figure*}

\begin{figure*}[!t]
\centering
\includegraphics[width=0.98\linewidth]{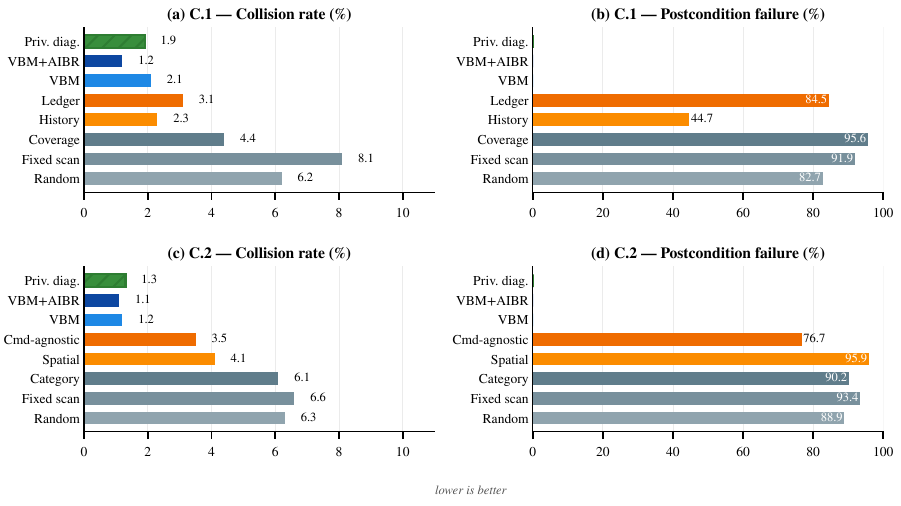}
\caption{\textbf{Closed-loop failure metrics at $H=5$.} Collision and postcondition failure are displayed separately. \textbf{Functional-OOD failure metrics at $H=3$.} Failure rates expose policies that observe plausible but command-irrelevant evidence. All rates use 1,500 episodes per method/protocol as denominator; the two failure metrics are nonexclusive.}
\label{fig:supp-failure-combined}
\label{fig:supp-c1-failure}
\label{fig:supp-c2-failure}
\end{figure*}

\begin{figure}[!t]
    \centering
    \includegraphics[width=\columnwidth]{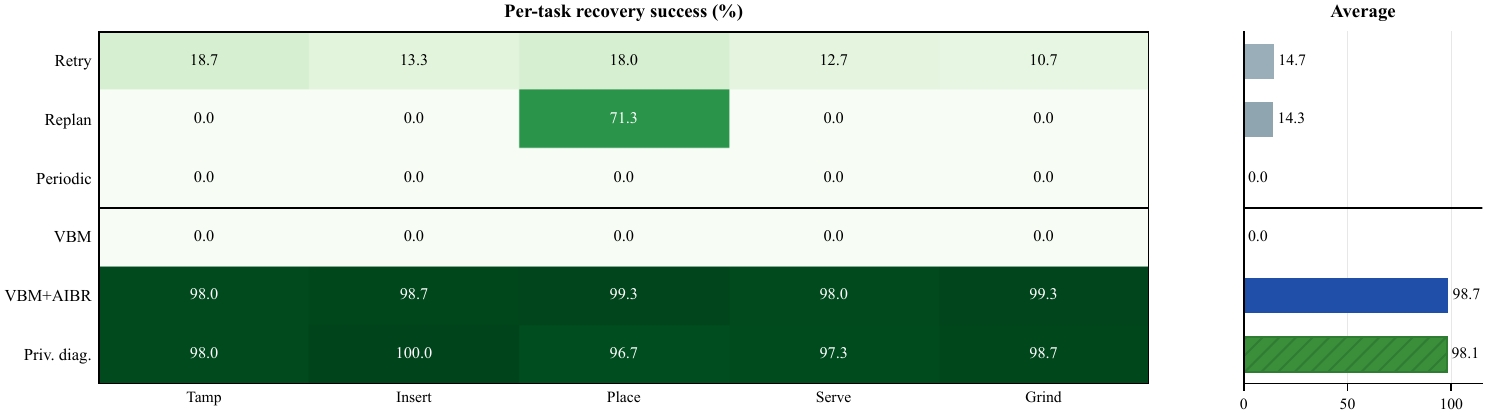}
    \caption{\textbf{Recovery after mismatch by task.} AIBR is evaluated as a repair diagnostic rather than the main source of primary-split gains. Denominators are 150 episodes per task/method and 750 per displayed method average.}
    \label{fig:supp-recovery-task}
\end{figure}

\begin{figure}[!t]
\centering
\includegraphics[width=\linewidth]{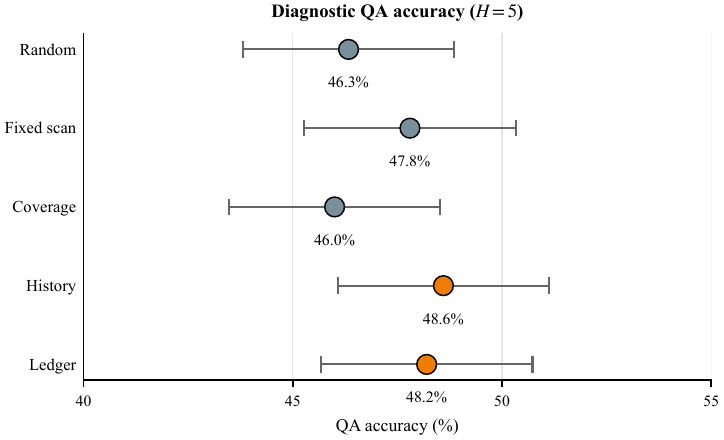}
\caption{\textbf{VLM diagnostic QA accuracy at $H=5$.} QA behavior is reported as a diagnostic of whether selected images contain answerable spatial-functional information; it is not treated as the main manipulation result. Each point reports \texttt{correct\_count}/$n$ with $n=1{,}500$; the normal-approximation 95\% binomial half-width is about 2.5 points. No random-answer chance level is defined; Random denotes random view selection.}
\label{fig:supp-qa-accuracy}
\end{figure}

Figures~\ref{fig:supp-task-combined}--\ref{fig:supp-qa-accuracy} respectively expose task-level success, nonexclusive failure modes, recovery by task, and diagnostic QA accuracy; none is used as a substitute for simulator postcondition success.

\section{Data Audit and Provenance}
\label{app:data-audit}

The regenerated figures use the frozen CSVs stored in \texttt{data/}. During audit, several transcription differences were found and corrected in the C.2 summaries, including the no-functional-fields row. In the final regenerated figures and tables, no-functional-fields success is 0.0\%; the 16.7 value refers to target-cue observation. The source package contains regenerated CSVs and verification records under \texttt{data/}, MuJoCo run and validation scripts under \texttt{scripts/}, and review records including \texttt{REPRODUCIBILITY.md}, \texttt{DATA\_CARD.md}, \texttt{ASSET\_ATTRIBUTION.md}, \texttt{COMPUTE\_REPORT.md}, and \texttt{APPENDIX\_QA\_CHECKLIST.md}. The files \texttt{DATA\_DIFFERENCES.md}, \texttt{figure\_catalog.csv}, \texttt{plotted\_values.csv}, and \texttt{verification.json} trace the plotted numbers and table values back to source summaries.

\paragraph{Asset and figure provenance.}
The two mechanism figures in the main paper are hybrid schematics depicting the coffee workstation, Drawer/Cabinet, and Microwave/Appliance with abstract modules and real MuJoCo RGB exemplars. They explain the benchmark interface and command-dependent evidence requirements; their annotations are not learned attention maps, segmentation masks, recorded selections, or quantitative outcome evidence. Quantitative plots in this supplement are generated from frozen CSV summaries; the mapping between plotted figures and source values is recorded in \texttt{figure\_catalog.csv} and \texttt{plotted\_values.csv}. Raw MuJoCo filmstrips document the simulator visual domain, while task success and failure claims are based on simulator-state postconditions.

\section{Reproducibility Record}
\label{sec:supp-reproducibility}

The review package is organized so that the main claims can be
regenerated from frozen records rather than from manually edited
numbers.
All experiment stages retain schedules, checkpoints, raw traces,
CSVs, generated tables, run logs, and SHA-256 manifests.
The package includes a reproducibility guide, data card,
asset-attribution note, CVPR-style compute report, and appendix QA checklist.
The two main MuJoCo output packages have SHA-256 prefixes
\texttt{045931cb445c} and \texttt{63011d469d}.
The canonical functional-OOD run validates with \texttt{ok:~true}, 55{,}500 episode rows, 55{,}500 schedule rows, and 55{,}500 trace lines.
The final chart package verifies 1{,}744 plotted/table source values after adding strict-clean VLM overall and per-task rows; it stores the plotted CSVs, figure catalog, and data-difference audit alongside the source.
The supplement includes seven verified 1{,}500-case strict-clean VLM rows, public RGB dataset package hashes, a hard $H{=}1$ stress summary derived from frozen episodes, a logged rollout trace figure generated from candidate-view scores and postcondition records, and a three-domain $H{=}3$ evaluation with protocol-revision audit.

\section{Limitations of the Supplementary Evidence}
\label{app:limitations}

The supplementary evidence strengthens auditability but does not over-broaden the paper's claim. The experiments evaluate simulated information selection and belief revision with frozen RGB candidate views, shared scripted skills, and hidden simulator-state postconditions. The direct VLM stress check and three-domain evaluation address two reviewer concerns--whether a prompted foundation model can be audited under a separate frozen full-preview contract and whether the formulation transfers beyond one coffee object family--but they do not demonstrate physical robot deployment, learned dexterous control, or household-scale generality. The main bootstrap table reports paired scene-group intervals over frozen cases, while budget-curve bands remain descriptive episode-level intervals. A future version should add independent environment seeds and real-robot trials where possible. Finally, diagnostic VLM QA remains a support check: it indicates whether selected images contain answerable spatial-functional information, but the main performance claim is closed-loop simulator postcondition success.

\section{Evidence Provenance and Release Chain}
\label{app:reconciliation}

The evidence package keeps complementary protocols explicit: the main paper's seven-model strict-clean experiment is supported by Tables~\ref{tab:strict-vlm-gate}--\ref{tab:strict-vlm-overall}, Fig.~\ref{fig:supp-vlm-comparison}, and Table~\ref{tab:strict-vlm-task-main}. Quantitative three-domain results appear in Tables~\ref{tab:p2-domain-protocol}--\ref{tab:p2-cross-domain}. Charged-view and ablation evidence appears in Tables~\ref{tab:supp-charged-views}--\ref{tab:supp-task-ablation} and Fig.~\ref{fig:supp-charged-views}. Content already present in the prior supplement is referenced in place rather than copied.

\begin{table}[!t]
\centering\scriptsize
\begin{tabularx}{\columnwidth}{p{0.30\columnwidth}X}
\tabletoprule
Benchmark component & Fixed and auditable element \\
\tablemidrule
Task-conditioned actionable belief & Evidence is judged by whether it makes the current command executable, not by generic scene QA alone. \\
Frozen RGB candidate views & Every method receives the same case-level candidate set; unselected endpoint renders and hidden labels are unavailable before action. \\
Hidden postcondition evaluator & Success, collision, and postcondition failure come from hidden simulator state after execution. \\
Shared scripted skills & All methods call the same low-level primitives; differences arise from evidence selection and belief revision. \\
Five-task case structure & Five tasks, 50 scene groups per task, two variants per group, and repeated inference/evaluation records. \\
C.1/C.2 protocols & C.1 tests closed-loop actionability; C.2 tests same-object functional-OOD command shifts. \\
Strict-clean VLM extension & Seven rows enter only after the 1,500-case endpoint/JSON/exact-$H=3$/membership/no-fallback gate. \\
Reproducibility package & Manifests, retry policy, response indexes, parsed selections, episode results, gates, audits, plotted values, and SHA256 records are packaged or referenced. \\
\tablebottomrule
\end{tabularx}
\caption{\textbf{Benchmark contribution checklist.} Reviewer-facing index of fixed and auditable components.}
\label{tab:contribution-checklist}
\end{table}

\paragraph{Evaluator API.}
The extension chain is
\begin{quote}\small
\texttt{manifest.csv} $\rightarrow$ \texttt{selected\_views.csv} $\rightarrow$\newline
\texttt{episode\_results.csv} $\rightarrow$ \texttt{strict\_gate.json} $\rightarrow$ summary tables.
\end{quote}

A new model uses the same command-plus-candidate-view contract, preserves raw responses and parsed selections, runs the unchanged postcondition evaluator, and passes the strict gate. A new domain additionally requires a simulator scene, Panda-compatible scripted skills, a hidden postcondition evaluator, a candidate-view renderer, and a frozen manifest with SHA256 records before a quantitative row is promoted.

\paragraph{Protocol-separation rules.}
The following records are never silently merged: (i) the 360-case/domain three-domain evaluation; (ii) the C.1 ablation-run full-method value 97.60\% and the 98.8\% primary closed-loop value; (iii) the fixed privileged diagnostic and an oracle upper bound; (iv) QA accuracy and manipulation postcondition success; and (v) old task-family bootstrap intervals and the final paired 500-case-cluster bootstrap intervals.

{\scriptsize
\paragraph{Source-record inventory.}
The reconciled numerical sources are \texttt{data/vlm\_strict1500\_overall.csv}, \texttt{data/vlm\_strict1500\_by\_task.csv}, \texttt{data/c1\_budget.csv}, \texttt{data/c2\_budget.csv}, \texttt{data/c1\_ablation.csv}, \texttt{data/c2\_ablation.csv}, \texttt{data/c1\_ablation\_tasks.csv}, and \texttt{data/c2\_ablation\_tasks.csv}. The editable main and supplement sources are packaged with these relative records; absolute machine paths are not required to reproduce the paper tables.\par
}

\end{document}